\documentclass[lettersize,journal]{IEEEtran}
\usepackage{amsmath,amsfonts}
\usepackage{algorithmic}
\usepackage{algorithm}
\usepackage{array}
\usepackage[caption=false,font=normalsize,labelfont=sf,textfont=sf]{subfig}
\usepackage{textcomp}
\usepackage{stfloats}
\usepackage{url}
\usepackage{graphicx}
\usepackage{cite}

\usepackage[table,xcdraw,dvipsnames,svgnames]{xcolor}
\definecolor{mygray}{HTML}{979C98}
\definecolor{myred}{HTML}{C12F51}
\definecolor{mygreen}{HTML}{A0A55C}
\definecolor{myblue}{HTML}{5698C3}
\definecolor{mybrown}{HTML}{975940}

\usepackage{amssymb}
\usepackage{colortbl} %
\usepackage{tabularx} %
\usepackage{booktabs}

\usepackage{multirow}
\usepackage{threeparttable}
\usepackage{makecell}

\usepackage{twemojis}
\usepackage{fontawesome5}
\usepackage{pifont}

\usepackage[shortlabels,inline]{enumitem}
\setlist[itemize]{noitemsep,leftmargin=*,topsep=0em}
\setlist[enumerate]{noitemsep,leftmargin=*,topsep=0em}

\usepackage[pagebackref=true,breaklinks=true,colorlinks=true]{hyperref}
\usepackage[capitalize]{cleveref}
\crefname{section}{Sec.}{Secs.}
\Crefname{section}{Section}{Sections}
\Crefname{table}{Table}{Tables}
\crefname{table}{Tab.}{Tabs.}

\newcommand{\none}{---}

\usepackage{xspace}
\newcommand{\parhead}[1]{\noindent\textbf{#1}\xspace}
\newcommand{\eg}{{\textit{e.g.},}\xspace}
\newcommand{\ie}{{\textit{i.e.},}\xspace}
\newcommand{\vs}{{\textit{vs.}}\xspace}
\newcommand{\mcnl}{\\}

\newcommand{\dataset}{\textsc{LaMoFCBench}\xspace}
\newcommand{\task}{LaMoFC\xspace}
\newcommand{\papertitle}{Towards Large Model Feature Coding}

\begin{document}

\title{\papertitle}

\author{%
    Youwei Pang\textsuperscript{1\,\ensuremath{\dagger}}
    Changsheng Gao\textsuperscript{1\,\ensuremath{\dagger}}
    Dong Liu\textsuperscript{2}
    Huchuan Lu\textsuperscript{3}
    Weisi Lin\textsuperscript{1\,\faEnvelope[regular]}
    \\
    \textsuperscript{1}NTU
    \textsuperscript{2}USTC
    \textsuperscript{3}DUT
    \\
    \thanks{\ensuremath{\dagger}: Equal contribution.}
    \thanks{\faEnvelope[regular]: Corresponding author.}%
}

\maketitle

\begin{abstract}
    Large models have delivered remarkable performance across a wide range of perception and generation tasks, yet practical deployment is increasingly constrained by computational and memory budgets, as well as privacy requirements.
    Split execution alleviates these constraints by partitioning computation across devices, but it inevitably introduces intensive transmission and storage of intermediate features.
    Unlike conventional feature coding for CNNs that typically targets homogeneous spatial activation maps, modern large models generate heterogeneous features with varying statistical distributions and compression tolerances, \eg multi-level/multi-modal representations and autoregressive context caches.
    These characteristics necessitate treating \emph{large model feature coding (\task)} as a fundamental system component and call for a systematic evaluation framework.
    In this paper, we present a comprehensive benchmark and evaluation framework for \task.
    We first build the feature dataset \dataset, covering diverse task requirements across 4 categories and 16 scenarios while integrating widely-adopted architectures and various split-computing settings.
    We then specify representative split points according to practical application scenarios to extract intermediate features, establishing a unified pipeline for fair and reproducible comparisons.
    Finally, we benchmark mainstream universal feature codecs, exposing the profound misalignment between existing coding paradigms and the heterogeneous nature of large model features.
    These findings reveal that \task demands a fundamental departure from existing paradigms, and \dataset provides the shared empirical foundation to drive this transition.
    The data and code will be available at \url{https://github.com/lartpang/LaMoFCBench}.
\end{abstract}

\begin{figure*}[t!]
    \centering
    \subfloat[Cloud-Centralized]{\label{fig:cloudcentric_application}
        \includegraphics[height=0.4\linewidth]{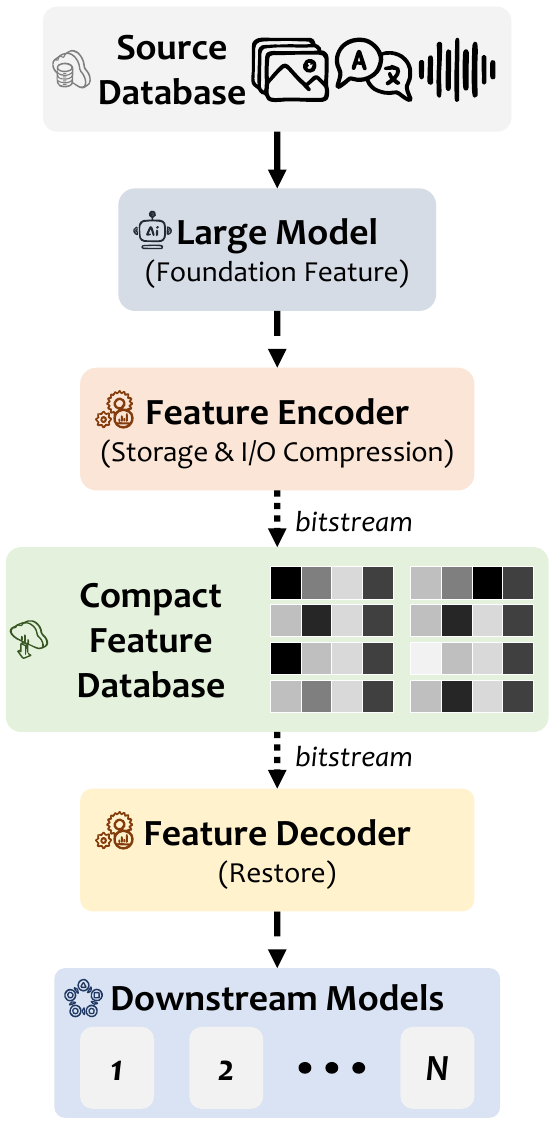}
    }
    \hspace{3ex}
    \subfloat[Cloud-Edge]{\label{fig:cloudedge_application}
        \includegraphics[height=0.4\linewidth]{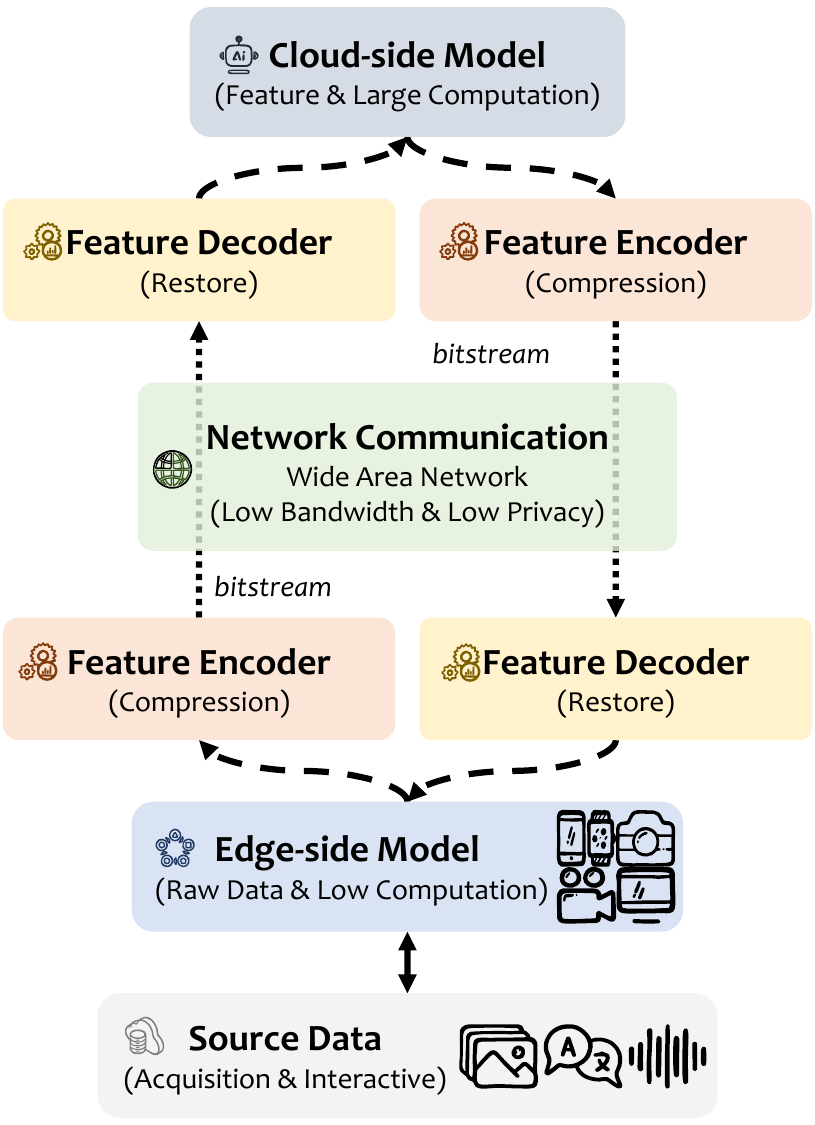}
    }
    \hspace{3ex}
    \subfloat[Edge-Edge]{\label{fig:edgeedge_application}
        \includegraphics[height=0.4\linewidth]{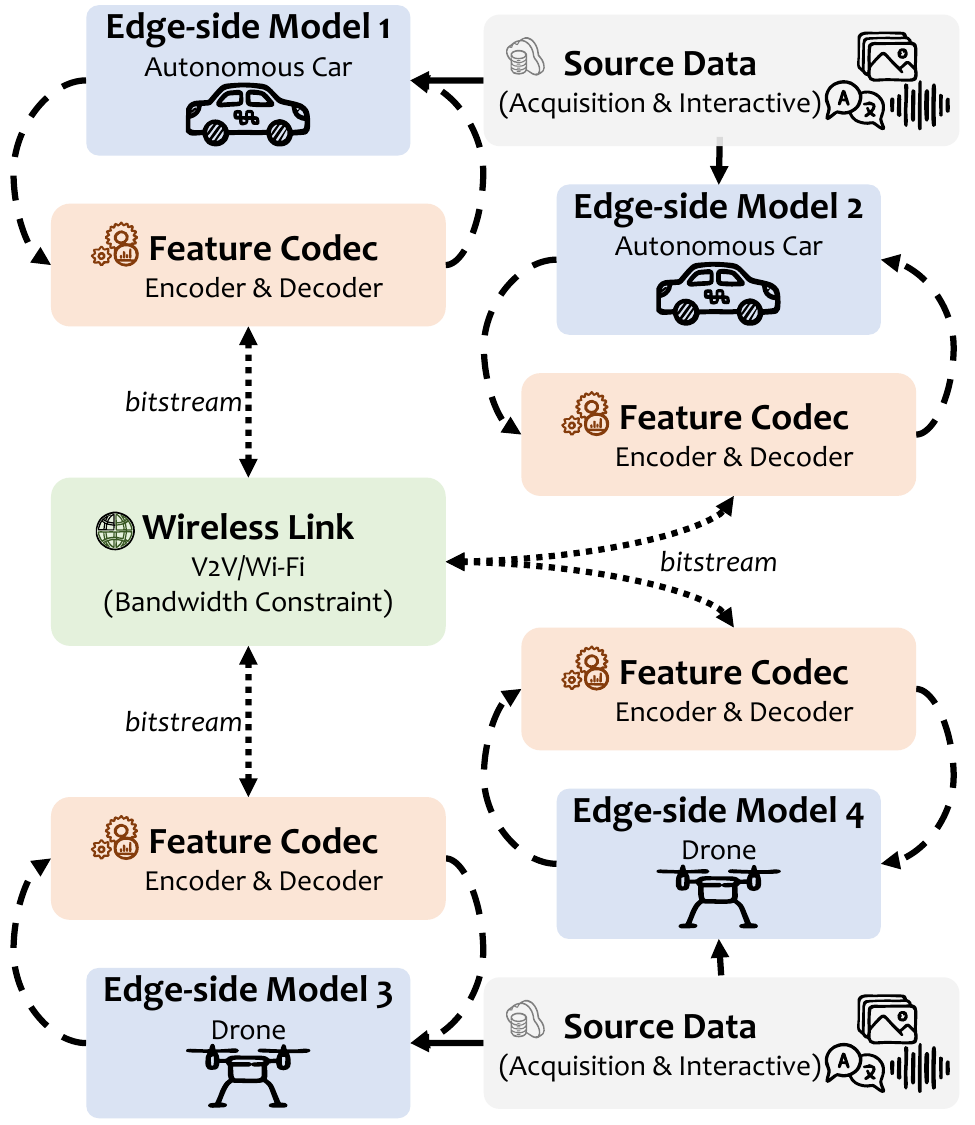}
    }
    \caption{Application scenarios for large model feature coding (\task).
        (a) (\cref{sec:cloud_centric_application}): Compresses large model features into a database to overcome storage and I/O bottlenecks during downstream training.
        (b) (\cref{sec:cloud_edge_application}): Transmits compressed intermediate features between edge and cloud devices, distributing computation, preserving data privacy and minimizing network bandwidth overhead.
        (c) (\cref{sec:edge_edge_application}): Compresses massive semantic features for efficient edge-to-edge exchange over constrained local networks, enabling low-latency decentralized collaboration.
    }
    \label{fig:fig_scenario}
    \vspace{-2ex}
\end{figure*}

\section{Introduction}
\label{sec:introduction}

\IEEEPARstart{L}{arge} models have recently driven remarkable progress across perception, reasoning, and generation.
However, scaling these models from laboratory success to widespread real-world deployment remains challenging due to the high cost of training and inference, the constrained compute and memory budgets of edge devices, and the growing need to keep sensitive data local~\cite{kaddour2023challenges}.
As a result, distributed deployment, particularly split-computing~\cite{vepakomma2018split,SplitComputingEarlyExiting-Survey}, has become a practical paradigm, \ie split execution across client-server or multi-platform systems~\cite{tian2022fedbert,vepakomma2018split,ye2024openfedllm,zheng2024safely,lepikhin2020gshard,friha2024llm}.
By partitioning computation among devices, it alleviates resource pressure on any single node and enables privacy-preserving use of on-device data while maintaining data security~\cite{ye2024openfedllm,chen2024federated}.
With large models continuing to expand in scale and diversify in modality and architecture, distributed deployment is poised to become a mainstream strategy.%
However, a fundamental bottleneck in this strategy is the exchange of intermediate information between model segments.
During forward propagation, the upstream segment produces intermediate representations that must be transmitted to the downstream segment to complete the forward pass.
At modern operating scales, directly transmitting these representations can dominate the end-to-end communication budget, increasing latency and energy consumption and enlarging the attack surface for privacy leakage.

Feature coding addresses the aforementioned bottleneck by compressing intermediate representations under semantics-preserving constraints, in spirit analogous to classical image coding pipelines but tailored to model-internal signals~\cite{chen2020toward,gao2024imofc,gao2024dmofc,wang2022towards}.
However, feature coding for large models departs substantially from the traditional setting and remains under-explored.
Large model deployments exhibit heterogeneity along three coupled axes.
\textbf{(1) Architecture and Modality.}
Representative architectures include the non-causal transformer \cite{Transformer-ViT} for vision, causal transformers \cite{Transformer} for language and audio, state-space model (SSM) \cite{LinearAttention-Mamba} for language, and diffusion-based generator \cite{Diffusion-DDPM}, each with distinct computational graphs, state evolution, and information correlation.
\textbf{(2) Intermediate Representation.}
The transmitted signal is often not a single activation tensor but a composition of heterogeneous representations.
Transformer-based models transmit prefill-stage hidden states and key/value caches.
State-space models additionally introduce SSM and convolution caches.
Diffusion pipelines require multi-encoder conditioning signals or denoised latents.
\textbf{(3) System- and Task-dependent Splitting.}
Split points vary with the device capability, latency constraints, and task requirements, yielding diverse data volumes, tensor shapes, and transmission rates.
These properties jointly imply that \textbf{large model feature coding (\task)} should be treated as a foundational problem for distributed deployment.
Accordingly, a reasonable evaluation framework for \task must adequately reflect this heterogeneity, covering diverse architectures and modalities, capturing practical intermediate representations across varied split depths, and assessing the practicality of codecs.
Despite its foundational importance, \task has received little attention to date.
Our conference version~\cite{FeatureCoding-LaMoFC} represents an early effort to explore this direction.
However, as an initial study, it remains limited in task scenarios,
modality types, and feature attributes, leaving the broader landscape of \task largely uncharted.

To bridge these gaps, we extend our preliminary exploration~\cite{FeatureCoding-LaMoFC} into a comprehensive benchmark and evaluation framework for \task.
We curate the \textbf{\dataset} dataset, spanning 4 categories and 16 scenarios, covering common vision, language, and audio understanding, as well as controllable text-to-image synthesis.
We further systematize split computing by introducing architecture-relevant split points and explicitly specifying multiple practical intermediate features (\eg multi-level representations, autoregressive context caches, multi-encoder conditioning representations, and denoised latents), enabling reasonable evaluation that better reflects real-world distributed deployment.
Crucially, we establish unified test conditions, including consistent precision-aware bitrate formulation, feature distortion measurement, and task-relevant evaluation pipelines, to ensure that future methods can be compared fairly under identical settings.
Beyond dataset construction, the practical relevance of feature coding hinges on reusability.
While learning-based codecs achieve strong performance when trained for a specific feature distribution, real-world deployments necessitate universal solutions~\cite{FeatureCoding-DT-UFC} capable of transferring across diverse scenarios with minimal re-training.
Moving beyond dataset construction, we shift our evaluation focus to mainstream, reusable coding schemes, investigating their behavior on heterogeneous features.
Recognizing that original codec implementations demand extensive hyperparameter tuning and task-specific retraining beyond the scope of a benchmark study, we specifically evaluate the universal variants~\cite{FeatureCoding-DT-UFC} of representative learning-based codecs~\cite{FeatureCoding-Hyperprior,FeatureCoding-ELIC}.%
\footnote{For brevity, subsequent sections use the original codec names to denote their respective universal variants.}
This strategy strictly aligns with the benchmark's core objective: \textbf{to evaluate the generalization capability and out-of-the-box performance of existing coding schemes on new large model features, rather than assessing their retraining potential}.
We report downstream task performance and feature reconstruction distortion, alongside codec-level analyses of generalizability and efficiency, enabling quantitative assessments aligned with practical deployment objectives.

In summary, our main contributions are as follows:
\begin{enumerate}
    \item \emph{Problem Formalization.}
          We formulate \textbf{large model feature coding (\task)} as a foundational problem for distributed deployment and systematically survey its diverse application scenarios.
          Through comprehensive analyses, we demonstrate the severe coding challenges posed by the heterogeneity of large model representations across diverse tasks and architectures.
    \item \emph{Comprehensive Dataset.}
          We meticulously construct the multi-modal \textbf{\dataset}, spanning 4 task categories and 16 scenarios.
          To ensure practical relevance, we specify representative split points to extract heterogeneous features, enabling rigorous split-aware evaluation.
    \item \emph{Standardized Evaluation.}
          We establish a unified feature-centric suite to standardize the testing protocol, incorporating comprehensive analyses across several critical dimensions: efficiency, distortion, efficacy, generalizability, and practicality, ensuring fair and reproducible comparisons.
    \item \emph{Codec Benchmarking.}
          We comprehensively benchmark representative universal codecs, exposing misalignments between existing image-centric coding paradigms and the nature of large model features.
          These findings yield actionable insights and delineate essential paradigm shifts for future feature-native coding research.
\end{enumerate}

The conference version~\cite{FeatureCoding-LaMoFC} of this work considers a limited set of tasks, models, feature data, and evaluation views.
In this paper, we substantially broaden the scope and enhance the realism across five key dimensions:
(1) %
\textbf{Task Scope}:
We expand task coverage to 16 scenarios across 4 categories, \ie common vision/language/audio understanding and controllable image synthesis.
(2) %
\textbf{Model Architecture}:
We update and broaden model coverage to more representative and diverse large model families, including non-causal and causal transformers, a state-space model, and a controllable diffusion generator.
(3) %
\textbf{Feature Types}:
We extend feature extraction from high-level tensors to a broader setting, encompassing diverse and practical split-computing representations, \eg early-layer hidden states, autoregressive context caches, and multi-encoder conditioning representations.
(4) %
\textbf{Evaluation Paradigm}:
We shift the focus from task-specific codec tuning to assessing reusable, universal coding schemes, prioritizing the generalization capability and out-of-the-box performance of existing technologies on new large model features rather than their retraining potential.
(5) %
\textbf{Experiment Analysis}:
We enrich experimental analyses by establishing a comprehensive measurement scheme, enabling an in-depth investigation into multiple critical dimensions, including intrinsic feature redundancy, coding efficiency, semantic distortion, overall efficacy, codec generalizability, and codec practicality.
Collectively, these extensions better reflect real application requirements and provide a more solid foundation for \task research.

\begin{table*}[t!]
    \centering
    \caption{Details of the proposed \textbf{\dataset}, organized by the source models, associated tasks (4 categories, 16 scenarios in total), source data (datasets, count, and metrics), and feature attributes (split point and shape). For features from language and audio models, $N$ in the shape denotes the length of the token sequence during the prefill stage, prior to autoregressive decoding.}
    \resizebox{\linewidth}{!}{\begin{threeparttable}
      \begin{tabular}{l|rrr|cc}
            \toprule
            \textbf{Model \& Task Setting}                              & \textbf{Source Dataset}                                                                                                                                                                                                                                                                                                                                   & \textbf{Count}         & \textbf{Metric}
                                                                        & \textbf{Split Point}                                                                                                                                                                                                                                                                                                                                      & \textbf{Feature Shape}                                      \\
            \midrule
            \multicolumn{3}{l}{DINOv3 (ViT7B)~\cite{DINOv3}}
                                                                        & \multicolumn{3}{r}{\textbf{\textit{Common Vision Understanding (CVU)}}}                                                                                                                                                                                                                                                                                                                                                 \\
            \midrule

            \quad Image Classification (Standard) with LinearHead       & ImageNet-Val~\cite{ImageNet1K}                                                                                                                                                                                                                                                                                                                            & 100                    & Accuracy ($\mathcal{A}$)$\uparrow$
                                                                        & \multirow{3}{*}{\makecell{Layer $10$ \mcnl Layer $40$}}
                                                                        & \multirow{3}{*}{\makecell{$1029 \times 4096$ \mcnl $1029 \times 4096$}}                                                                                                                                                                                                                                                                                                                                                 \\
            \quad Image Classification (Robustness) with LinearHead     & ImageNet-A~\cite{ImageNet-A}                                                                                                                                                                                                                                                                                                                              & 100                    & Accuracy ($\mathcal{A}$)$\uparrow$
                                                                        &                                                                                                                                                                                                                                                                                                                                                           &                                                             \\
            \quad Image Classification (Generalization) with LinearHead & ImageNet-R~\cite{ImageNet-R}                                                                                                                                                                                                                                                                                                                              & 100                    & Accuracy ($\mathcal{A}$)$\uparrow$
                                                                        &                                                                                                                                                                                                                                                                                                                                                           &                                                             \\
            \midrule

            \multirow{1}{*}{\quad Semantic Segmentation with Mask2FormerHead~\cite{Mask2Former}}
                                                                        & ADE20K-Val~\cite{ADE20K}                                                                                                                                                                                                                                                                                                                                  & 100                    & mIoU$\uparrow$
                                                                        & \multirow{2}{*}{Layers ($10, 20, 30, 40$)}
                                                                        & $4 \times 2 \times 3141 \times 4096$\tnote{\ding{192}}                                                                                                                                                                                                                                                                                                                                                                  \\
            \multirow{1}{*}{\quad Depth Estimation with DPTHead~\cite{DPT}}
                                                                        & NYUDepthV2-Test~\cite{NYUDepthV2}                                                                                                                                                                                                                                                                                                                         & 100                    & RMSE$\downarrow$
                                                                        &
                                                                        & $4 \times 2 \times 3077 \times 4096$\tnote{\ding{192}}                                                                                                                                                                                                                                                                                                                                                                  \\

            \midrule
            \multicolumn{3}{l}{Qwen3 (8B)~\cite{Qwen3} and FalconMamba (7B)~\cite{FalconMamba}}
                                                                        & \multicolumn{3}{r}{\textbf{\textit{Common Language Understanding (CLU)}}}                                                                                                                                                                                                                                                                                                                                               \\
            \midrule

            \quad Mathematical Reasoning                                & GSM8K~\cite{GSM8K}                                                                                                                                                                                                                                                                                                                                        & 100                    & Accuracy ($\mathcal{A}$)$\uparrow$
                                                                        & \multirow{5}{*}{$\begin{matrix} \text{Layer }5 \\ \text{(Prefill Stage)}\end{matrix}$}
                                                                        & \multirow{5}{*}{$\begin{aligned} \text{Qwen3 Hidden State: }& N \times 3584 \mcnl \text{Qwen3 Key/Value Cache: }& 5 \times 4 \times N \times 128, 5 \times 4 \times N \times 128 \mcnl \text{FalconMamba Hidden State: }& N \times 4096 \mcnl \text{FalconMamba SSM/Convolution Cache: }& 5 \times 8192 \times 16, 5 \times 8192 \times 4 \end{aligned}$}                                                               \\
            \quad Knowledge Evaluation                                  & ArcChallenge~\cite{AI2Arc}                                                                                                                                                                                                                                                                                                                                & 100                    & Accuracy ($\mathcal{A}$)$\uparrow$
                                                                        &                                                                                                                                                                                                                                                                                                                                                           &                                                             \\
            \quad Truthfulness Evaluation                               & TruthfulQA~\cite{TruthfulQA}                                                                                                                                                                                                                                                                                                                              & 100                    & Accuracy ($\mathcal{A}$)$\uparrow$
                                                                        &                                                                                                                                                                                                                                                                                                                                                           &                                                             \\
            \quad Commonsense Inference                                 & Hellaswag~\cite{HellaSwag}                                                                                                                                                                                                                                                                                                                                & 100                    & Accuracy ($\mathcal{A}$)$\uparrow$
                                                                        &                                                                                                                                                                                                                                                                                                                                                           &                                                             \\
            \quad Ambiguity Resolution                                  & Winogrande~\cite{WinoGrande}                                                                                                                                                                                                                                                                                                                              & 100                    & Accuracy ($\mathcal{A}$)$\uparrow$
                                                                        &                                                                                                                                                                                                                                                                                                                                                           &                                                             \\

            \midrule
            \multicolumn{3}{l}{KimiAudio (7B)~\cite{KimiAudio}}
                                                                        & \multicolumn{3}{r}{\textbf{\textit{Common Audio Understanding (CAU)}}}                                                                                                                                                                                                                                                                                                                                                  \\
            \midrule

            \quad Automatic Speech Recognition (Clean)                  & LibriSpeech-Test-Clean~\cite{Librispeech}                                                                                                                                                                                                                                                                                                                 & 100                    & WER$\downarrow$
                                                                        & \multirow{5}{*}{$\begin{matrix} \text{Layer }5 \mcnl \text{(Prefill Stage)}\end{matrix}$}
                                                                        & \multirow{5}{*}{$\begin{aligned} \text{Hidden State: }& N \times 3584 \mcnl \text{Key Cache: }& 5 \times 4 \times N \times 128 \mcnl \text{Value Cache: }& 5 \times 4 \times N \times 128 \end{aligned}$}                                                                                                                                                                                                               \\
            \quad Automatic Speech Recognition (Noisy)                  & LibriSpeech-Test-Other~\cite{Librispeech}                                                                                                                                                                                                                                                                                                                 & 100                    & WER$\downarrow$
                                                                        &                                                                                                                                                                                                                                                                                                                                                           &                                                             \\
            \quad Audio Adversarial Defense                             & AdvBench~\cite{VoiceBench}\tnote{\ding{193}}                                                                                                                                                                                                                                                                                                              & 100                    & Accuracy ($\mathcal{A}$)$\uparrow$
                                                                        &                                                                                                                                                                                                                                                                                                                                                           &                                                             \\
            \quad Spoken Scientific Reasoning                           & OpenBookQA~\cite{VoiceBench}\tnote{\ding{193}}                                                                                                                                                                                                                                                                                                            & 100                    & Accuracy ($\mathcal{A}$)$\uparrow$
                                                                        &                                                                                                                                                                                                                                                                                                                                                           &                                                             \\
            \quad Dialect-Robust Question Answering                     & SD-QA~\cite{VoiceBench}\tnote{\ding{193}}                                                                                                                                                                                                                                                                                                                 & 100                    & Accuracy ($\mathcal{A}$)$\uparrow$
                                                                        &                                                                                                                                                                                                                                                                                                                                                           &                                                             \\
            \midrule

            \multicolumn{3}{l}{SD3.5\tnote{\ding{194}}~~(8B)~\cite{StableDiffusion3.5}}
                                                                        & \multicolumn{3}{r}{\textbf{\textit{Controllable Text-to-Image Synthesis (CTTI)}}}                                                                                                                                                                                                                                                                                                                                       \\
            \midrule

            \quad Controllable Synthesis with ControlNet~\cite{ControlNet}
                                                                        & \makecell[c]{COCO2017-Val~\cite{MSCOCO} \mcnl (Caption~\cite{MSCOCOCpation}+Edge)\tnote{\ding{195}}}                                                                                                                                                                                                                                                      & 100                    & FID$\downarrow$
                                                                        & \makecell[l]{$\begin{matrix} \text{Text Encoders} \mcnl \text{(CLIP-L/G, T5-XXL)} \mcnl \text{VAE Encoder} \mcnl \text{Visual Latent} \end{matrix}$}
                                                                        & $\begin{aligned} \text{Text Embedding 1: }& 768, 77 \times 768 \mcnl \text{Text Embedding 2: }& 1280, 77 \times 1280 \mcnl \text{Text Embedding 3: }& 77 \times 4096 \mcnl \text{Image Embedding: }& 32 \times 128 \times 128 \mcnl \text{Latent: }& 16 \times 128 \times 128 \end{aligned}$                                                                                                                            \\
            \bottomrule
      \end{tabular}
      \begin{tablenotes}
            \item[\ding{192}] Following the original implementations, two test-time augmented copies are derived from the input images.
            \item[\ding{193}] These datasets are collected and released in VoiceBench~\cite{VoiceBench}.
            \item[\ding{194}] \textbf{For brevity, we refer to Stable Diffusion 3.5 as SD3.5 throughout this paper.}
            \item[\ding{195}] We generate the edge image by applying the Canny filter to the original image corresponding to the caption~\cite{MSCOCOCpation}.
      \end{tablenotes}
\end{threeparttable}
}
    \label{tab:table_summary}
    \vspace{-2ex}
\end{table*}

\section{Background and Motivation}
\label{sec:background_and_motivation}

\subsection{Evolution from Visual Coding to \task}
\label{sec:feature_coding_evolution}

Feature coding originated within the visual coding for machines framework~\cite{yang2021video,duan2020video}, serving as a machine-centric counterpart to traditional visual coding.
Unlike visual coding~\cite{tian2023nonsemantics,zhang2024all,gao2023towards,lu2024preprocessing,tian2024smc,mao2024perceptual,sheng2024vnvc,li2024object,li2024ustc}, which aims to reconstruct original visual data for human perception, feature coding encodes intermediate representations in models tailored for machine applications.
Current feature coding research has been predominantly developed for CNNs~\cite{li2023attention,suzuki2022deep,kim2023end,liu2023learnt,cai2022high,ma2024feature,gao2024rethinking,CNNFeatureCoding-EmergingStandards,CNNFeatureCoding-GlobalStatisticsPreservation,CNNFeatureCoding-MSFIBA,CNNFeatureCoding-NewVVCProfiles,CNNFeatureCoding-NextGenerationConsumerExperience,CNNFeatureCoding-ScalableMachineVision,CNNFeatureCoding-StereoImageCodingJointFeatureCompression}.
As large models shift dramatically towards transformers~\cite{Transformer,Transformer-ViT,DINOv3} and state-space models~\cite{LinearAttention-Mamba}, the structure and form of features have fundamentally changed.
We define large model feature coding (\task) as the compression of heterogeneous internal signals from diverse large models, ranging from patch tokens to autoregressive context caches and conditioning latents, specifically to facilitate their distributed deployment.

\subsection{Scaling Laws and Deployment Challenges}
\label{sec:scaling_laws}

The need for \task arises from the intersection of scaling laws~\cite{ScalingLaws-Hestness,ScalingLaws-Kaplan,ScalingLaws-Chinchilla} and the rigid constraints of real-world deployment.
Scaling laws highlight that model performance improves predictably as parameter count and training data increase, but such scaling incurs substantial computational and storage costs.
To bridge the gap between these massive models and resource-constrained edge devices, distributed deployment~\cite{ye2024openfedllm,zheng2024safely,friha2024llm}, particularly split computing~\cite{vepakomma2018split,SplitComputingEarlyExiting-Survey}, has become essential.
Crucially, privacy protection requirements~\cite{chen2024federated,lyu2024privacy,yan2024protecting} in client-facing services further constrain deployment.
As a practical approach, split computing partitions the model across the client and server, enabling privacy-preserving inference by keeping raw sensitive data on-device~\cite{ye2024openfedllm}.
However, model scaling also amplifies a new bottleneck for split computing, \ie the communication overhead caused by the rapid growth in intermediate feature volume.
In practice, the transmitted feature volume can exceed that of the raw input, making inter-device bandwidth the dominant bottleneck.
Current research often overlooks the transmission costs associated with this feature exchange.
To address this gap, we propose encoding features into compact bitstreams, making \task a prerequisite for efficient large model deployments.

\subsection{Large Model Compression}
\label{sec:large_model_compression}

The massive computational and memory footprint of large models has driven extensive research into model compression~\cite{Survey-LLMCompression}.
While weight compression~\cite{Quantization-GPTQ,Quantization-AWQ} reduces the static memory of model parameters, it operates on stored weights rather than runtime signals, and is therefore orthogonal to \task.
A more closely related line of research targets the intermediate activations produced during inference, particularly the key/value caches in transformers, to alleviate the memory wall during long-context generation.
These methods typically employ token dropping \cite{EfficientInference-H2O,KVCacheCompression-Scissorhands} or low-bit quantization \cite{KVCacheCompression-KVQuant} to reduce cache size.
Despite the similarity in operating on runtime activations, this line of work differs from \task in objective, mechanism, and design space.
Existing methods are designed to overcome the memory bound on a single GPU or node, often relying on hardware-friendly quantization or model-specific pruning strategies.
In contrast, \task explicitly targets the transmission bound in distributed deployments.
It requires universal, distribution-aware coding paradigms that compress heterogeneous activations across diverse modalities into bandwidth-efficient bitstreams for transmission across wide-area or resource-constrained links, while preserving semantic utility for downstream tasks.

\subsection{Critical Misalignments in Current Research}
\label{sec:research_gaps}

Despite the emerging need for \task, current research exhibits three misalignments with the practical realities of large models, motivating our comprehensive benchmark.

\subsubsection{Map-Centric \vs Token-Centric}
The most significant gap lies in the target architecture.
Existing methods~\cite{li2023attention,suzuki2022deep,kim2023end,liu2023learnt,cai2022high,ma2024feature,gao2024rethinking,CNNFeatureCoding-EmergingStandards,CNNFeatureCoding-GlobalStatisticsPreservation,CNNFeatureCoding-MSFIBA,CNNFeatureCoding-NewVVCProfiles,CNNFeatureCoding-NextGenerationConsumerExperience,CNNFeatureCoding-ScalableMachineVision,CNNFeatureCoding-StereoImageCodingJointFeatureCompression} are mainly optimized for CNN feature maps, which typically exhibit spatial locality and more stationary statistics amenable to local spatial transforms~\cite{FeatureCoding-LaMoFC}.
In contrast, modern large models utilize token-based representations that exhibit architecture-specific fingerprints, as shown in~\cref{fig:histogram_cdf_distribution}.
For example, vision transformer features exhibit a depth-driven evolution, shifting from concentrated activations in shallow layers to broad, multi-peaked distributions with expanding dynamic ranges in deeper layers.
Moreover, transformer or state-space models introduce functional differences across context caches.
For instance, in transformer-based Qwen3~\cite{Qwen3}, key caches manifest high variance and a multi-peaked distribution, whereas value caches exhibit smoother and single-peaked distributions concentrated within a narrower dynamic range.%
This statistical heterogeneity poses a major challenge for conventional codecs typically optimized for homogeneous representations.

\subsubsection{Discrimination \vs Generation}
Current studies mainly focus on discriminative tasks, such as classification and segmentation~\cite{choi2021latent,feng2022image,zhang2021MSFC,yan2021SSSIC,chen2024end,misra2022video}.
However, the current AI wave is driven by Generative AI, encompassing autoregressive language/audio understanding models~\cite{Qwen3,FalconMamba,KimiAudio} and diffusion-based synthesis models~\cite{StableDiffusion3.5}.
Tasks like text-to-image synthesis impose stricter fidelity requirements, as minor distortions in the conditioning representations (\eg text prompts or control maps~\cite{ControlNet}) can lead to catastrophic semantic drift in the generated content.
The absence of evaluation on generative tasks creates a critical empirical gap.

\subsubsection{Vision-Only \vs Multi-Modal}
The scope of existing feature coding~\cite{choi2021latent,feng2022image,zhang2021MSFC,yan2021SSSIC,chen2024end,misra2022video,li2023attention,suzuki2022deep,kim2023end,liu2023learnt,cai2022high,ma2024feature,gao2024rethinking,CNNFeatureCoding-EmergingStandards,CNNFeatureCoding-GlobalStatisticsPreservation,CNNFeatureCoding-MSFIBA,CNNFeatureCoding-NewVVCProfiles,CNNFeatureCoding-NextGenerationConsumerExperience,CNNFeatureCoding-ScalableMachineVision,CNNFeatureCoding-StereoImageCodingJointFeatureCompression} remains confined to visual feature maps, neglecting the growing significance of features generated from other modalities.
However, modern AI applications are increasingly multi-modal.
As systems evolve to understand language and audio signals, the transmitted payload diversifies beyond spatial tensors, relying on sequence embeddings and autoregressive context caches.
These non-visual representations possess different statistical behaviors and distortion tolerances compared to visual data.
The absence of a comprehensive multi-modal feature benchmark prevents the community from developing unified coding schemes capable of seamlessly handling the diverse modality types in modern AI systems.

\section{Application Scenarios}
\label{sec:application_scenarios}

In real-world deployments, \task plays a pivotal role by alleviating storage and transmission bottlenecks, while providing a privacy-enhancing interface that operates on intermediate features.
In~\cref{fig:fig_scenario}, we categorize \task applications based on data-flow topology and resource constraints into: cloud-centralized, cloud-edge, and edge-edge.

\subsection{Cloud-Centralized Application}
\label{sec:cloud_centric_application}

In the era of large models, a prevailing paradigm is ``pre-training once, fine-tuning everywhere''.
Large models, such as transformer backbones~\cite{DINOv3}, are used to extract features from massive datasets.
These features then serve as input for training lightweight heads~\cite{CloudCentricTraining} tailored to downstream tasks like image classification, semantic segmentation~\cite{Mask2Former} or depth estimation~\cite{DPT}.
Since re-running large model inference for every downstream task is computationally prohibitive, features are typically extracted once and reused, making compact and efficient feature storage essential.
However, this strategy shifts the bottleneck from computation (FLOPs) to storage and input/output (I/O).
Storing raw floating-point features for millions of images can consume petabytes of disk space, creating severe latency issues due to disk-to-memory bandwidth limitations during training.
\task addresses this challenge by compressing raw features into a compact feature database.
As shown in~\cref{fig:cloudcentric_application}, the large model processes the source data once, and the resulting features are encoded and archived.
For subsequent tasks, the training pipeline reads and decodes these compact bitstreams directly.
This approach significantly reduces storage costs and alleviates I/O congestion.
Since the computational overhead of decoding features is negligible compared to the latency of reading massive uncompressed files from disk, feature coding effectively accelerates the end-to-end training throughput for diverse downstream applications.

\subsection{Cloud-Edge Application}
\label{sec:cloud_edge_application}

The interaction between resource-constrained edge devices and powerful cloud servers faces two primary challenges: (i) privacy regulations that discourage centralizing raw user data, limiting the availability of training data, and (ii) the limited uplink/downlink bandwidth of wide-area networks (WAN).
\task provides a unified feature-space coding layer to transmit model features efficiently, making split collaboration practical at scale.
For \textbf{distributed training}, sensitive data (\eg personal records or photos) can remain on-device.
Instead of uploading raw data, the edge computes the initial model layers and uploads intermediate features to the cloud to finish the remaining forward propagation, while gradients are returned for backpropagation.
\task is critical here to compress both features and gradients into compact bitstreams to fit tight link budgets.
Furthermore, operating in feature space avoids direct exposure of raw inputs, and compression/quantization further suppresses fine-grained signals, offering a lightweight privacy-enhancing effect that complements system-level protections.
For \textbf{distributed inference}, feature coding reduces transmission latency while enabling privacy-oriented designs.
This is particularly relevant to interactive applications such as controllable text-to-image synthesis involved in this paper.
As shown in~\cref{fig:cloudedge_application}, the uplink carries compressed control inputs.
These may be raw condition inputs (\eg text prompts or compressed edge maps) or their on-device encoded embeddings/features to guide cloud-side generation.
\task provides a unified way to minimize payload while retaining controllability.
On the downlink, the cloud may complete the entire generation and return the final image (bandwidth-efficient with mature image codecs).
Alternatively, it may also return only features and offload the final decoding to the edge.
The latter reduces cloud-side handling of the final pixels and becomes bandwidth-feasible when latents are compressed, especially under higher resolution, higher precision, or richer intermediate representations.
\task makes this design practical by compactly encoding feature-space payloads for efficient transmission.

\subsection{Edge-Edge Application}
\label{sec:edge_edge_application}

In edge collaborative scenarios lacking stable cloud connectivity (\eg autonomous driving fleets~\cite{Edge2Edge-V2VFormer,Edge2Edge-V2VFormer++}, advanced drone networks~\cite{Edge2Edge-DHD}), computing devices must collaborate via local wireless links to extend their sensing coverage and execute complex joint decision-making.
Compared with human-centric communication in pixel, text, or audio space, machine-to-machine collaboration often relies on exchanging semantic representations for efficient perception and coordination.
To handle complex real-world dynamics, large models are rapidly introduced into this paradigm~\cite{Edge2Edge-DINO-CoDT,Device2EdgeServer-ToFC}.
However, the massive high-dimensional intermediate features generated by large models far exceed the volume of conventional sensory data, leading to severe network congestion and latency.
As shown in~\cref{fig:edgeedge_application}, \task can serve as a key technology to break this communication bottleneck by compressing features into compact bitstreams.
This naturally requires that codecs preserve the internal semantics required for complex downstream reasoning.
Moreover, \task provides lightweight privacy protection~\cite{PrivacyInV2V} by limiting precision and discarding fine-grained details.
Looking forward, fully decentralized large model collaboration remains a largely unexplored frontier.
As computing capability grows and \task matures, future edge devices will be able to sustain increasingly frequent and dense representation exchange, enabling more natural and seamless intelligent interaction among distributed edge units.

\begin{figure*}[t!]
    \centering
    \renewcommand{\tabularxcolumn}[1]{m{#1}}
    \begin{tabularx}{\linewidth}{@{} c @{\hspace{0em}} X @{}}
        \rotatebox[origin=c]{90}{\scriptsize \textbf{CVU (DINOv3)}}      &
        \includegraphics[width=\linewidth]{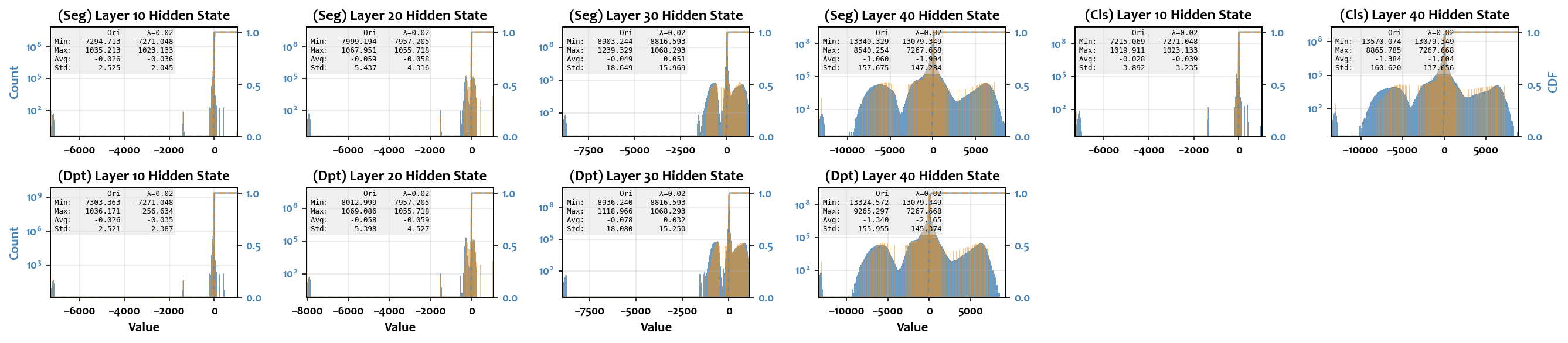}
        \\ \vspace{-1ex}
        \rotatebox[origin=c]{90}{\scriptsize \textbf{CLU (Qwen3)}}       &
        \includegraphics[width=\linewidth]{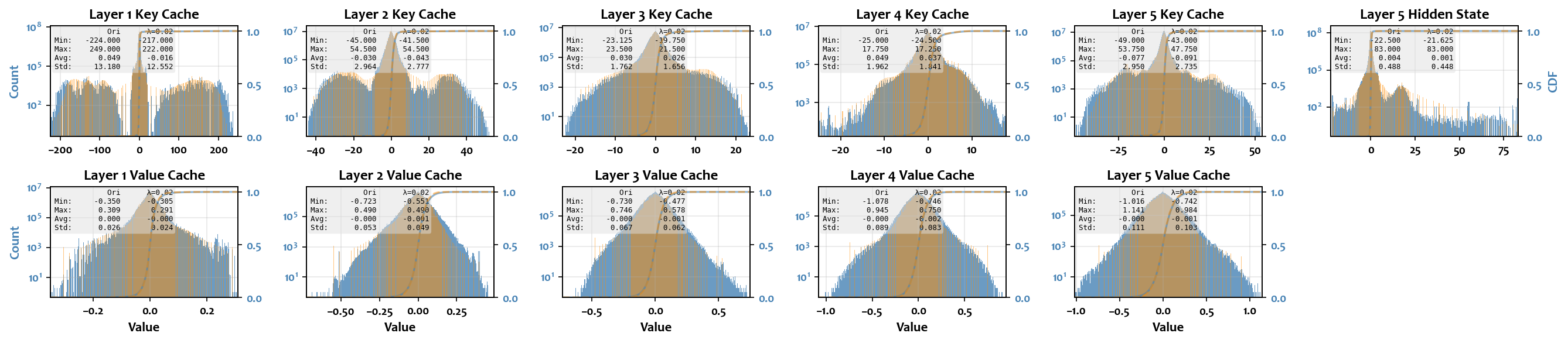}
        \\ \vspace{-1ex}
        \rotatebox[origin=c]{90}{\scriptsize \textbf{CLU (FalconMamba)}} &
        \includegraphics[width=\linewidth]{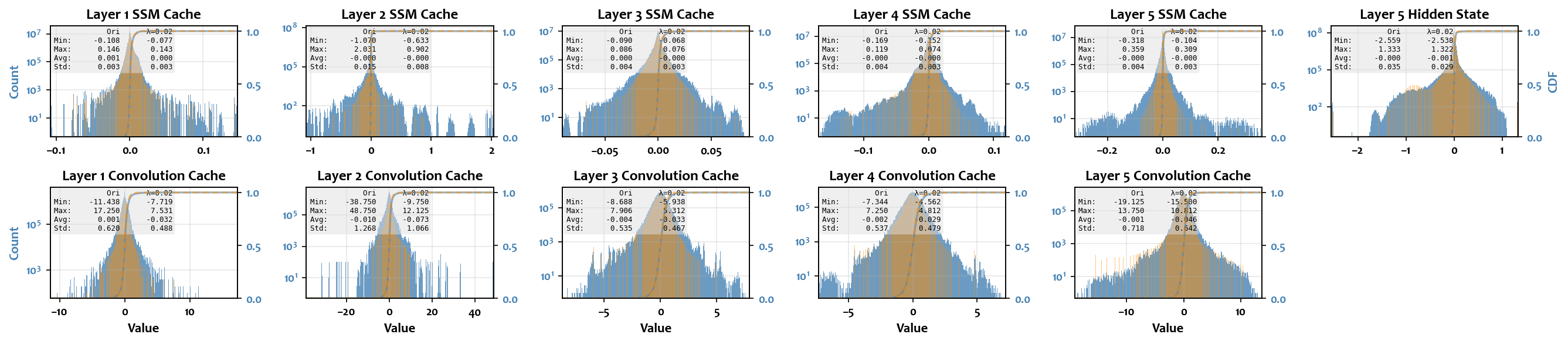}
        \\ \vspace{-1ex}
        \rotatebox[origin=c]{90}{\scriptsize \textbf{CAU (KimiAudio)}}   &
        \includegraphics[width=\linewidth]{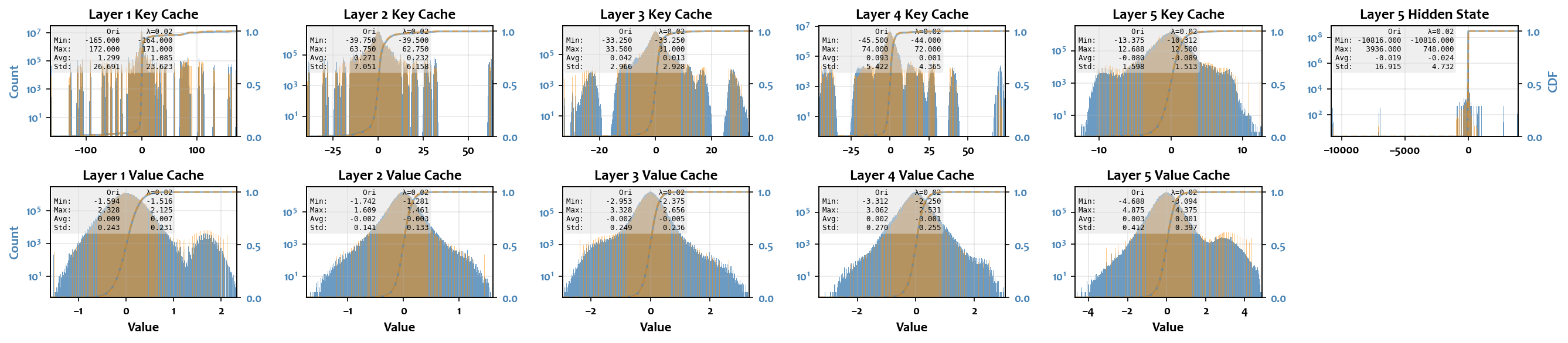}
        \\ \vspace{-1ex}
        \rotatebox[origin=c]{90}{\scriptsize \textbf{CTTI (SD3.5)}}      &
        \includegraphics[width=\linewidth]{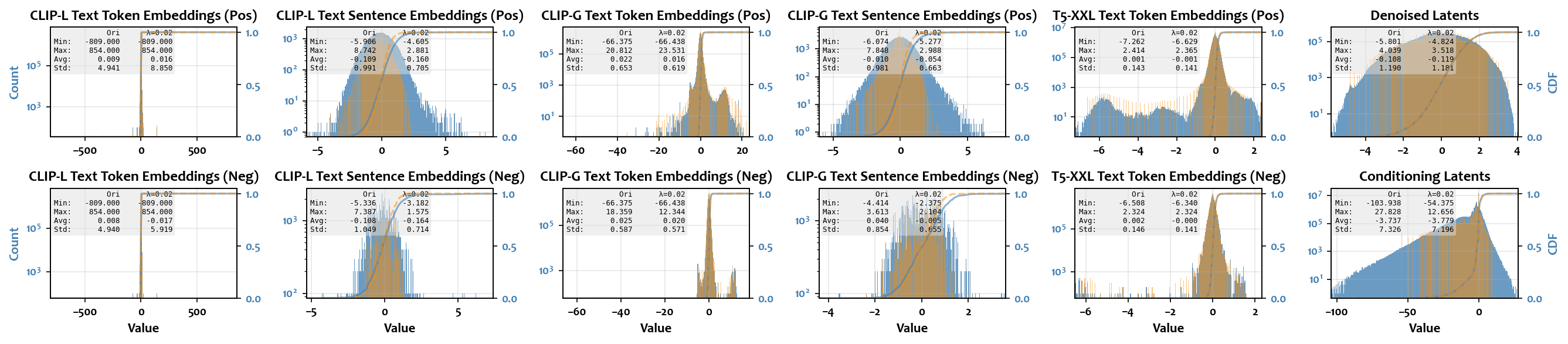}
    \end{tabularx}
    \caption{Histogram and cumulative distribution function (CDF) curve comparisons between the \textcolor{DarkOrange}{reconstructed features} from the ELIC baseline ($\lambda$=0.02) and \textcolor{SteelBlue}{original features} across different data subsets. These features exhibit diverse distributions, demonstrating the necessity and value of our dataset.}
    \label{fig:histogram_cdf_distribution}
\end{figure*}

\section{Dataset Construction}
\label{sec:dataset_construction}

To guarantee the dataset's representativeness and support the long-term research, we curate \dataset, considering three key aspects, as summarized in~\cref{tab:table_summary}.

\subsection{Model and Task Selection}
\label{sec:model_and_task_selection}

Given the rapid proliferation of large models, it is impractical to cover every architecture.
To ensure broad representativeness, we select widely adopted models in the 7--8B parameter range for each task category:
DINOv3~\cite{DINOv3} for common vision understanding,
Qwen3~\cite{Qwen3} and FalconMamba~\cite{FalconMamba} for common language understanding,
KimiAudio~\cite{KimiAudio} for common audio understanding,
and SD3.5~\cite{StableDiffusion3.5} equipped with ControlNet~\cite{ControlNet} for controllable text-to-image synthesis.
These selections span both discriminative and generative paradigms, and align with prevailing scalable sequence-modeling trends in large model development (transformers~\cite{DINOv3,Qwen3,KimiAudio,StableDiffusion3.5} and state-space models~\cite{FalconMamba}).
Together, they cover visual, textual, auditory, and cross-modal settings:
DINOv3 processes visual inputs in five vision scenarios,
Qwen3 and FalconMamba model text in five text scenarios,
KimiAudio reasons over auditory signals conditioned on text in five audio scenarios,
and SD3.5 generates images from text with additional visual conditioning.
Collectively, this suite provides a compact yet comprehensive basis for evaluating feature coding across diverse model and task types.

\subsection{Data Source and Metrics}
\label{sec:data_source_metrics}

We utilize the chosen large models to extract feature subsets.

\parhead{Common Vision Understanding (CVU).}
We utilize DINOv3 to extract visual features across varying granularities.
For image classification, we build a comprehensive evaluation set from ImageNet-Val~\cite{ImageNet1K} (standard), ImageNet-A~\cite{ImageNet-A} (robustness), and ImageNet-R~\cite{ImageNet-R} (generalization), using accuracy as the primary metric.
For ImageNet-Val, we select 100 correctly predicted samples from unique classes as originally done~\cite{FeatureCoding-LaMoFC}.
For the other two sets, where accurate classes are few, we select the top 100 highest-confidence correct instances across distinct classes.
To assess the performance in dense prediction tasks, we source data from ADE20K-Val~\cite{ADE20K} for semantic segmentation and NYUDepthV2-Test~\cite{NYUDepthV2} for depth estimation, measured by mIoU and RMSE, respectively.
From ADE20K-Val (150 classes), we select 100 samples by prioritizing class coverage and high mIoU.
For depth estimation, we evenly allocate top-performing samples across scene categories and select the remainder based on the lowest RMSE.

\parhead{Common Language Understanding (CLU).}
To evaluate coding on sequential reasoning, we employ both Qwen3 and FalconMamba, focusing on hidden states and context caches in the prefill stage.
We assess mathematical reasoning using GSM8K \cite{GSM8K} and evaluate comprehensive language understanding by aggregating samples from four diverse benchmarks: ArcChallenge \cite{AI2Arc} (knowledge), TruthfulQA \cite{TruthfulQA} (truthfulness), Hellaswag \cite{HellaSwag} (commonsense), and Winogrande \cite{WinoGrande} (ambiguity resolution).
Using accuracy as the metric, we curate the evaluation set by selecting the 100 longest correctly predicted instances from each dataset.

\parhead{Common Audio Understanding (CAU).}
Leveraging KimiAudio, we consider both signal-level transcription and semantic understanding.
For automatic speech recognition (ASR), we evaluate on the clean and noisy subsets of LibriSpeech~\cite{Librispeech} using the word error rate (WER), and sample the 100 longest correctly transcribed instances from each subset.
For audio question answering (AQA), we use VoiceBench~\cite{VoiceBench} subsets, \ie AdvBench (audio adversarial defense), OpenBookQA (spoken scientific reasoning), and SD-QA (dialect-robust QA), measured by accuracy.
We select the 100 longest correctly answered instances per subset.

\parhead{Controllable Text-to-Image Synthesis (CTTI).}
We focus on controllable generation using ControlNet-enhanced SD3.5.
Using COCO2017-Val \cite{MSCOCO}, we retain the longest caption per image, encode captions with CLIP \cite{VLM-CLIP}, cluster the embeddings into 100 groups via K-means, and select the real sample nearest to each cluster center.
For each selected image, we derive a Canny edge map as an additional visual condition.
We use FID~\cite{Metric-FID} to quantify the distribution shift of generated images relative to those conditioned on the original features.

\subsection{Feature Characteristics}
\label{sec:feature_characteristics}

To accommodate diverse deployment requirements, we carefully deliberate on the split point selection tailored to the computational characteristics and task demands of each architecture.
Rather than restricting extraction to the deepest layer, we shift split points towards shallower stages that better reflect practical split-computing scenarios for each architecture and task.
\cref{tab:table_summary} summarizes the selected split points and the resulting feature information.

\parhead{Feed-forward Representation.}
For DINOv3, we adopt a hierarchical feature extraction protocol to support workload redistribution at different task complexities.
For image classification, we use a dual-point setting by extracting features from Layers 10 and 40 of the vision transformer.
For semantic segmentation and depth estimation tasks, we aggregate multi-level outputs from Layers (10, 20, 30, 40).
These settings facilitate two computational paradigms:
(1) a lightweight edge mode that transmits early features to offload most representation computation to the cloud server, reducing on-device latency;
and (2) a high-separability mode where the upstream segment completes encoding, allowing the downstream segment to focus on decoding prediction using deep features for classification or multi-level features for dense scene understanding.

\parhead{Autoregressive Modeling.}
Autoregressive language and audio models typically operate under the standard prefill-decode paradigm.
The prefill stage processes the full context in a single pass and produces a burst of intermediate activations along with the initial context caches, whereas the decode stage updates them token-by-token in a latency-critical loop.
This workload attribute makes the prefill stage more amenable to feature coding.
Compressing the one-shot intermediate features enables an efficient feature-space interface, while avoiding per-token overhead and potential error accumulation during decoding.
Accordingly, we set the split point at the output of an early layer (\ie Layer 5) during prefill, and encode the resulting features for transmission and reuse by downstream layers.
This design is applied to Qwen3, FalconMamba, and KimiAudio.
Crucially, the extracted features include not only hidden states but also context caches required for autoregressive generation, \ie the key/value cache for transformer-based models (Qwen3, KimiAudio) and the SSM/convolution cache for the Mamba-based architecture (FalconMamba).

\parhead{Controllable Synthesis.}
For text-to-image synthesis, we use ControlNet-equipped SD3.5 to study coding of heterogeneous multi-modal conditions.
We place split points at the outputs of the conditioning encoders to capture multi-modal representations before diffusion denoising begins.
Following the configuration in our conference version~\cite{FeatureCoding-LaMoFC}, we additionally place a split point before the VAE decoder to enable redistribution of generative workloads.
The resulting feature set is highly heterogeneous, including three text embeddings from the triple-encoder design (CLIP-L\&G~\cite{VLM-CLIP} and T5-XXL~\cite{LanguageModel-T5}), conditioning latents extracted from Canny edge maps, and denoised latents before the VAE decoder.
This setting serves as a rigorous testbed for assessing whether essential generative priors can be preserved under high compression ratios.

\section{Feature Data Analysis}
\label{sec:feature_data_analysis}

In this section, we comprehensively analyze feature statistical behaviors across diverse architectures, highlighting the distinct challenges they pose for feature coding.

\subsection{Distribution Analysis}
\label{sec:distribution_analysis}

\cref{fig:histogram_cdf_distribution} presents the feature distributions (histogram) and cumulative distribution functions (CDF).
Our dataset reveals that feature distributions are not merely non-stationary but exhibit unique, architecture-specific fingerprints.

\parhead{CVU.}
Regarding DINOv3, we investigate the intrinsic layer-wise evolution of feature statistics under a frozen backbone setting~\cite{DINOv3}.
In shallower layers (\ie Layer 10), feature distributions are concentrated within a narrow range and exhibit noticeable asymmetry, with activations predominantly residing in the negative domain.
As the network deepens from Layer 20 to Layer 40, the feature statistics undergo a progressive expansion in dynamic range accompanied by increased distributional complexity.
The emergence of broad, multi-peaked distributions spanning both positive and negative extremes indicates a substantial energy reallocation.

\parhead{CLU.}
The feature distributions in large language models are dictated by their internal functional components.
For Qwen3, the key and value caches display divergent statistics.
The value cache tends to follow a concentrated distribution with a narrow dynamic range.
In contrast, the key cache exhibits higher variance and more heterogeneous statistics, supporting the differentiation required for attention queries.
In FalconMamba, the SSM cache is highly peaked and range-constrained, while the convolution cache is smoother and broader in range.
This highlights that text features are not a monolithic category.
Their statistical properties are intrinsically tied to the underlying sequence modeling mechanism (attention \vs state space).

\parhead{CAU and CTTI.}
KimiAudio and SD3.5 introduce more diversity in feature statistics.
KimiAudio features demonstrate a unique form of discretization.
Specifically, the key cache (\eg Layer 1) presents a distinct comb-like distribution with separated value clusters across a wide dynamic range.
It reflects a structured discretization in which values concentrate at isolated clusters, rather than vanishing towards zero as in conventional sparsity.
In SD3.5, we observe a hierarchical dichotomy within the conditioning space itself.
Compared to text token embeddings that are highly spiky and heavy-tailed, global sentence embeddings and denoised latents follow smoother distributions with lower variance.

\parhead{Discussion.}
These observations reveal a complex and heterogeneous statistical landscape.
As \task lacks established standards and baselines, our analysis provides an early characterization of large model feature behaviors, spanning depth-dependent distribution shifts in ViTs, functional heterogeneity across context caches, and discretized representations in multi-modal generation.
By capturing these architecture-dependent variations, the proposed dataset offers a shared foundation for the community.
It delineates the emerging problem space and supplies the data needed to catalyze the development of adaptive, architecture-aware feature coding schemes.

\begin{table*}[t!]
    \centering
    \begin{minipage}[t]{0.245\linewidth}
        \centering
        \caption{Feature redundancy analysis for CVU (DINOv3).}
        \label{tab:dinov3_feature_analysis}
        \resizebox{!}{9em}{

\begin{tabular}{c|cccc}
    \toprule
    Layer & $\rho_h$ & $\rho_v$ & $G_\text{DCT}$ & $C_\text{DCT}$ \\
    \midrule
    \multicolumn{5}{r}{Image Classification}                      \\
    \midrule
    9     & +0.000   & +0.139   & 0.682          & 0.509          \\
    39    & +0.000   & +0.764   & 0.982          & 0.486          \\
    \midrule
    \multicolumn{5}{r}{Semantic Segmentation}                     \\
    \midrule
    9     & +0.000   & +0.255   & 0.743          & 0.489          \\
    19    & +0.001   & +0.494   & 0.920          & 0.423          \\
    29    & +0.000   & +0.617   & 0.962          & 0.409          \\
    39    & +0.000   & +0.857   & 0.989          & 0.400          \\
    \midrule
    \multicolumn{5}{r}{Depth Estimation}                          \\
    \midrule
    9     & +0.000   & +0.262   & 0.736          & 0.492          \\
    19    & +0.001   & +0.513   & 0.918          & 0.426          \\
    29    & -0.001   & +0.636   & 0.960          & 0.412          \\
    39    & +0.001   & +0.864   & 0.987          & 0.403          \\
    \bottomrule
\end{tabular}
}
    \end{minipage}
    \begin{minipage}[t]{0.245\linewidth}
        \centering
        \caption{Feature redundancy analysis for CLU (Qwen3).}
        \label{tab:qwen3_feature_analysis}
        \resizebox{!}{9em}{

\begin{tabular}{c|cccc}
    \toprule
    Layer & $\rho_h$ & $\rho_v$ & $G_\text{DCT}$ & $C_\text{DCT}$ \\
    \midrule
    \multicolumn{5}{r}{Hidden State}                              \\
    \midrule
    4     & +0.002   & +0.118   & 0.823          & 0.499          \\
    \midrule
    \multicolumn{5}{r}{Key Cache}                                 \\
    \midrule
    0     & +0.003   & +0.107   & 0.985          & 0.491          \\
    1     & -0.012   & +0.243   & 0.950          & 0.497          \\
    2     & +0.018   & +0.285   & 0.878          & 0.487          \\
    3     & +0.065   & +0.326   & 0.880          & 0.470          \\
    4     & +0.018   & +0.296   & 0.935          & 0.487          \\
    \midrule
    \multicolumn{5}{r}{Value Cache}                               \\
    \midrule
    0     & -0.012   & -0.022   & 0.682          & 0.511          \\
    1     & -0.004   & +0.010   & 0.668          & 0.509          \\
    2     & -0.006   & +0.072   & 0.669          & 0.509          \\
    3     & -0.002   & +0.133   & 0.691          & 0.505          \\
    4     & -0.005   & +0.098   & 0.667          & 0.508          \\
    \bottomrule
\end{tabular}
}
    \end{minipage}
    \begin{minipage}[t]{0.245\linewidth}
        \centering
        \caption{Feature redundancy analysis for CLU (FalconMamba).}
        \label{tab:falconmamba_feature_analysis}
        \resizebox{!}{9em}{

\begin{tabular}{c|cccc}
    \toprule
    Layer & $\rho_h$ & $\rho_v$ & $G_\text{DCT}$ & $C_\text{DCT}$ \\
    \midrule
    \multicolumn{5}{r}{Hidden State}                             \\
    \midrule
    4     & +0.000   & +0.022   & 0.706          & 0.497          \\
    \midrule
    \multicolumn{5}{r}{Convolution Cache}                               \\
    \midrule
    0     & -0.388   & -0.004   & 0.658          & 0.502          \\
    1     & -0.357   & -0.008   & 0.716          & 0.503          \\
    2     & -0.365   & +0.005   & 0.644          & 0.497          \\
    3     & -0.379   & +0.001   & 0.649          & 0.500          \\
    4     & -0.353   & +0.001   & 0.663          & 0.499          \\
    \midrule
    \multicolumn{5}{r}{SSM Cache}                                 \\
    \midrule
    0     & -0.098   & +0.004   & 0.760          & 0.413          \\
    1     & -0.102   & +0.000   & 0.705          & 0.495          \\
    2     & +0.045   & -0.002   & 0.753          & 0.493          \\
    3     & +0.108   & -0.008   & 0.732          & 0.495          \\
    4     & -0.261   & -0.005   & 0.717          & 0.500          \\
    \bottomrule
\end{tabular}
}
    \end{minipage}
    \begin{minipage}[t]{0.245\linewidth}
        \centering
        \caption{Feature redundancy analysis for CAU (KimiAudio).}
        \label{tab:kimiaudio_feature_analysis}
        \resizebox{!}{9em}{

\begin{tabular}{c|cccc}
    \toprule
    Layer & $\rho_h$ & $\rho_v$ & $G_\text{DCT}$ & $C_\text{DCT}$ \\
    \midrule
    \multicolumn{5}{r}{Hidden State}                             \\
    \midrule
    4     & +0.002   & +0.308   & 0.698          & 0.500          \\
    \midrule
    \multicolumn{5}{r}{Key Cache}                                \\
    \midrule
    0     & +0.086   & +0.497   & 0.997          & 0.429          \\
    1     & +0.033   & +0.608   & 0.992          & 0.448          \\
    2     & +0.006   & +0.632   & 0.974          & 0.466          \\
    3     & +0.057   & +0.594   & 0.988          & 0.449          \\
    4     & +0.009   & +0.628   & 0.887          & 0.468          \\
    \midrule
    \multicolumn{5}{r}{Value Cache}                               \\
    \midrule
    0     & +0.002   & +0.400   & 0.747          & 0.490          \\
    1     & -0.009   & +0.380   & 0.719          & 0.498          \\
    2     & -0.001   & +0.460   & 0.741          & 0.489          \\
    3     & -0.004   & +0.499   & 0.735          & 0.490          \\
    4     & -0.003   & +0.538   & 0.804          & 0.483          \\
    \bottomrule
\end{tabular}
}
    \end{minipage}
\end{table*}
\begin{table}[t!]
    \centering
    \caption{Feature redundancy analysis for CTTI (SD3.5).}
    \label{tab:sd35cond_feature_analysis}
    \resizebox{\linewidth}{!}{

\begin{tabular}{l|l|ccc|cccc}
    \toprule
    \multirow{2}{*}{Layer}        &
    \multirow{2}{*}{Feature Data} &
    \multicolumn{3}{c|}{Sentence Feature (1D)}
                                  &
    \multicolumn{4}{c}{Token Feature (2D)}                                                                                                                    \\
                                  &                      & $\rho_h$ & $G_\text{DCT}$ & $C_\text{DCT}$ & $\rho_h$ & $\rho_v$ & $G_\text{DCT}$ & $C_\text{DCT}$ \\
    \midrule
    \multicolumn{9}{r}{Multi-Modal Conditioning Signals}                                                                                                      \\
    \midrule
    CLIP-L                        & Text Embedding (Pos) & -0.015   & 0.635          & 0.501          & -0.006   & +0.382   & 0.703          & 0.502          \\
    CLIP-L                        & Text Embedding (Neg) & +0.023   & 0.662          & 0.487          & -0.039   & +0.600   & 0.704          & 0.502          \\
    CLIP-G                        & Text Embedding (Pos) & -0.006   & 0.638          & 0.502          & -0.013   & +0.402   & 0.860          & 0.504          \\
    CLIP-G                        & Text Embedding (Neg) & -0.011   & 0.633          & 0.505          & -0.032   & +0.665   & 0.909          & 0.510          \\
    T5-XXL                        & Text Embedding (Pos) & \none    & \none          & \none          & +0.003   & +0.589   & 0.935          & 0.499          \\
    T5-XXL                        & Text Embedding (Neg) & \none    & \none          & \none          & +0.005   & +0.900   & 0.978          & 0.497          \\
    VAE Encoder                   & Conditioning Latents & \none    & \none          & \none          & +0.803   & +0.901   & 0.992          & 0.058          \\
    \midrule
    \multicolumn{9}{r}{Denoised Latents}                                                                                                                      \\
    \midrule
    MM-DiT                        & Denoised Latents     & \none    & \none          & \none          & +0.921   & +0.929   & 0.981          & 0.040          \\
    \bottomrule
\end{tabular}
}
\end{table}

\subsection{Redundancy Analysis}
\label{sec:redundancy_analysis}

We investigate the redundancy of the individual features%
\footnote{Given the feature diversity, the packing strategy is flexible. For clarity, we adopt the simplest \emph{independent packing} scheme, where each feature is packed separately. More details can be found in~\cref{sec:feature_coding_pipeline}.}
using two complementary metrics:
(1) spatial correlation, quantified by the Pearson coefficients ($\rho$) between adjacent elements along the horizontal ($\rho_h$) and vertical ($\rho_v$) axes,
and (2) energy distribution, measured by the Gini coefficient ($G_{\text{DCT}}$) and the normalized centroid ($C_{\text{DCT}}$) in the DCT domain.%
\footnote{A higher $\rho$ indicates stronger local smoothness (higher spatial redundancy), whereas $\rho \approx 0$ suggests weak linear dependence between adjacent elements.
    A larger $G_{\text{DCT}}$ indicates stronger energy compaction, \ie higher sparsity.
    A smaller $C_{\text{DCT}}$ indicates energy concentrated in low frequencies, while a larger value implies a shift towards higher-frequency components.
    See supplementary material for details.}
The statistics are summarized in~\cref{tab:dinov3_feature_analysis,tab:qwen3_feature_analysis,tab:falconmamba_feature_analysis,tab:kimiaudio_feature_analysis,tab:sd35cond_feature_analysis}.

\parhead{Spatial Correlation.}
In our packing configuration, the $v$-axis of DINOv3, Qwen3, and KimiAudio corresponds to the token sequence, while the $h$-axis corresponds to the feature-channel dimension.
From~\cref{tab:dinov3_feature_analysis,tab:kimiaudio_feature_analysis}, we observe consistently stronger sequence-wise correlation than channel-wise correlation.
This evidence shows substantial continuity along the sequence dimension, whereas feature channels are largely decorrelated (\ie near-zero $\rho_h$).
It suggests that semantic content varies smoothly across tokens (high $\rho_v$), while information is distributed across approximately independent channels (low $\rho_h$).
FalconMamba illustrates how correlation patterns depend on component design.
For convolution caches, we map the inner dimension to the $v$-axis and the temporal window (local receptive field) to the $h$-axis.
The statistics indicate weak correlation across the inner dimension (low $\rho_v$) but noticeable correlation along the temporal window (moderate $\rho_h$).
In contrast, SSM caches exhibit weak correlation along both the inner dimension ($v$) and the state dimension ($h$), reflecting the decorrelated nature of state-space variables.
SD3.5 exhibits hybrid correlation structures.
For conditioning and denoised latents, where $h$ and $v$ correspond to spatial width and height, we observe strong and approximately isotropic spatial correlation.
For token embeddings, the correlation follows the transformer pattern discussed above: high token-wise correlation ($v$) but negligible channel-wise correlation ($h$).
For sentence-level global embeddings, $\rho_h$ remains near zero, indicating a weakly correlated channel structure.

\parhead{Energy Distribution.}
We further study frequency-domain structure by computing $G_{\text{DCT}}$ and $C_{\text{DCT}}$ to quantify spectral concentration and frequency bias.
Results in~\cref{tab:dinov3_feature_analysis,tab:qwen3_feature_analysis,tab:falconmamba_feature_analysis,tab:kimiaudio_feature_analysis,tab:sd35cond_feature_analysis} show that many large model features exhibit non-trivial spectral concentration.
Specifically, for DINOv3, Qwen3, and FalconMamba, $G_{\text{DCT}}$ typically lies in the range of $0.6$-$1.0$, indicating that energy is dominated by a subset of frequency components.
However, the corresponding $C_{\text{DCT}}$ values suggest that this dominance is not confined to low frequencies, consistent with a flattened spectrum.
It implies that semantic information may be carried by frequency components beyond the low-frequency band.
By contrast, features retaining spatial structure (\eg certain latent representations) tend to exhibit stronger low-frequency concentration (higher $G_{\text{DCT}}$ and lower $C_{\text{DCT}}$) than token-based semantic representations.

\parhead{Discussion.}
Across the selected models, the correlation and spectral statistics reveal that redundancy characteristics in large model features are highly heterogeneous across architectures and internal components.
Therefore, feature codecs should explicitly account for the dimension-specific dependencies characterized by our dataset.

\begin{figure*}[t!]
    \centering
    \subfloat[Classification (Layer 10)]{\includegraphics[width=0.245\linewidth]{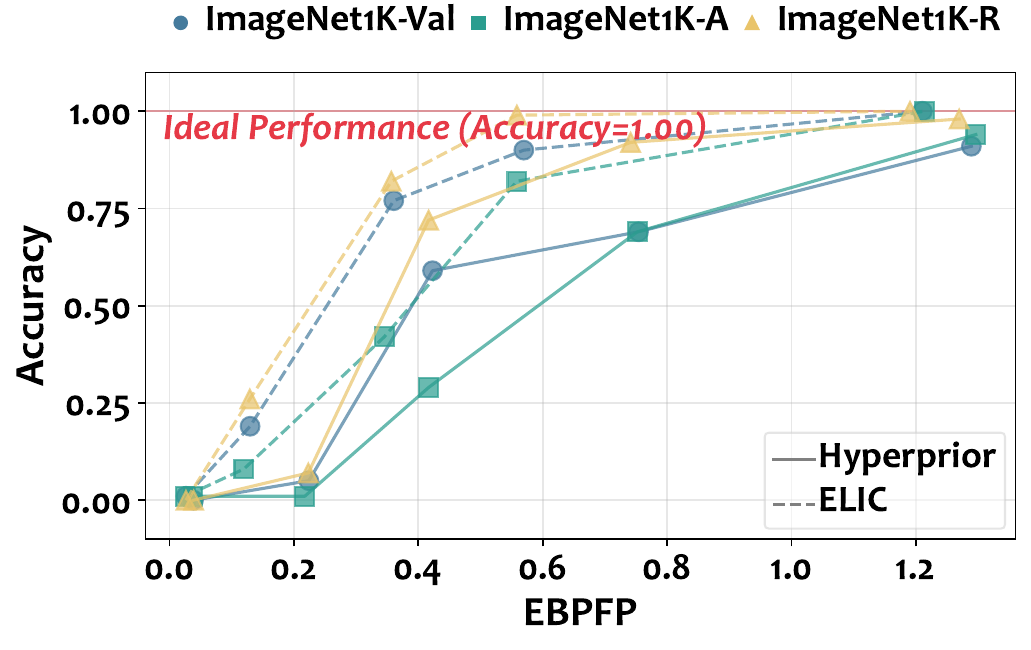}}
    \subfloat[Classification (Layer 40)]{\includegraphics[width=0.245\linewidth]{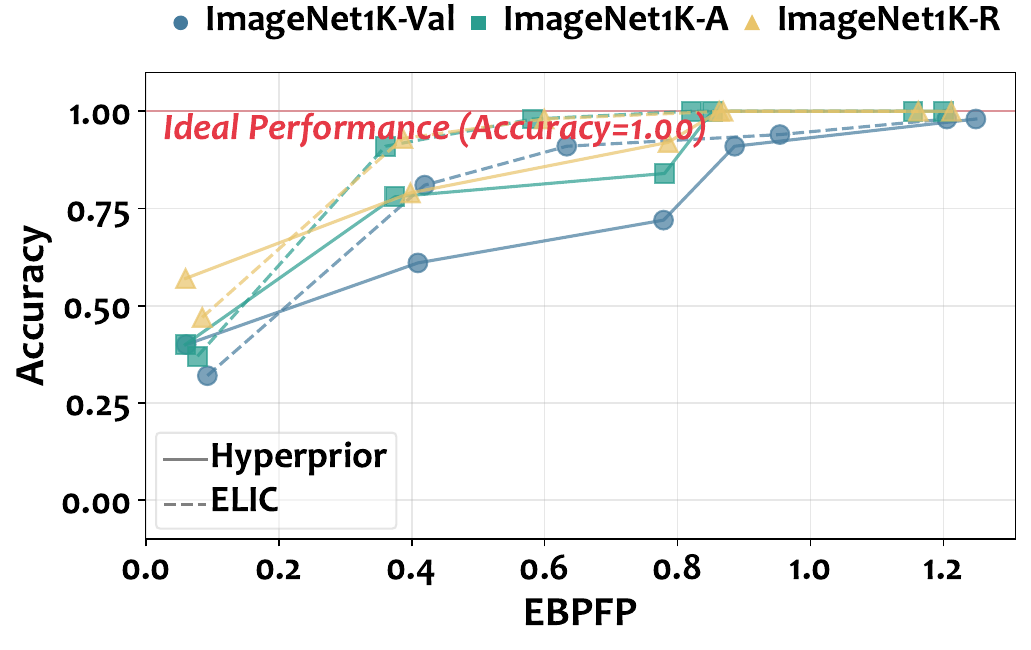}}
    \subfloat[Semantic Segmentation]{\includegraphics[width=0.245\linewidth]{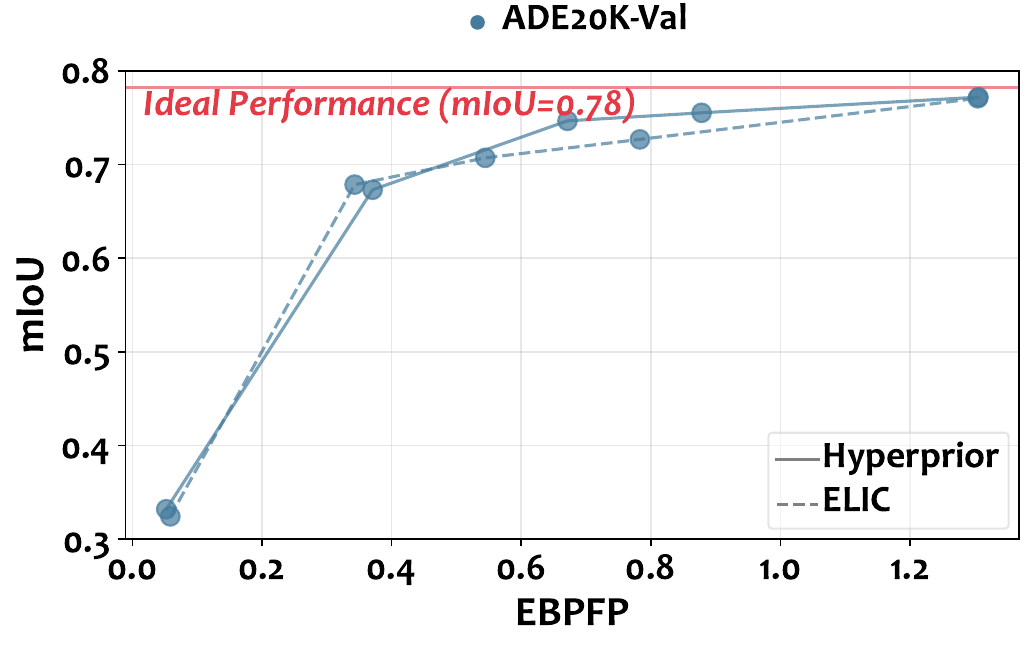}}
    \subfloat[Depth Estimation]{\includegraphics[width=0.245\linewidth]{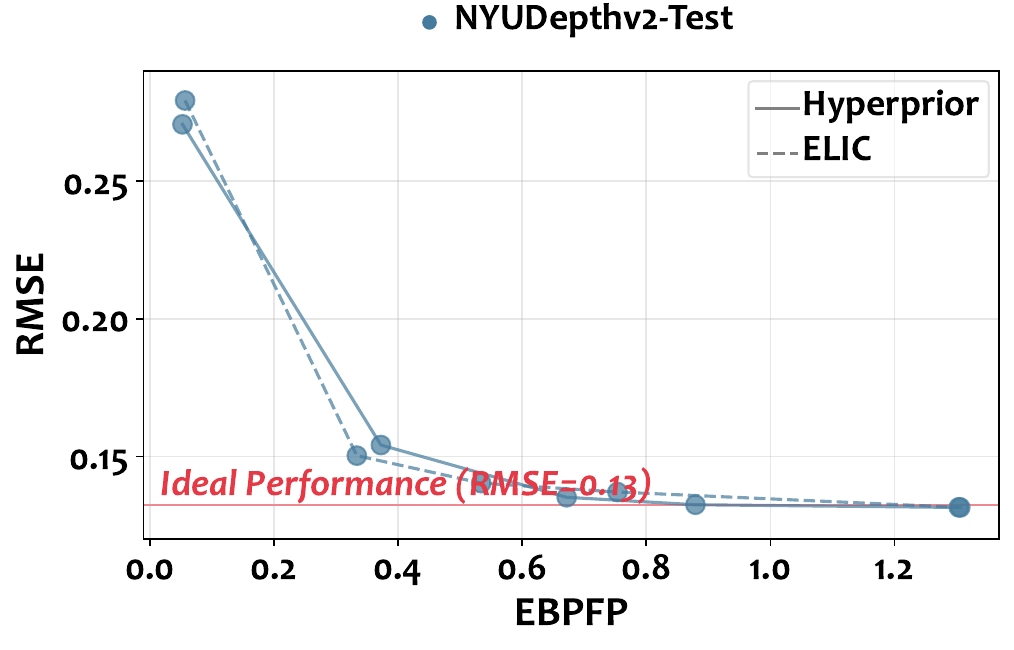}}
    \caption{Rate-performance curves for CVU (DNIOv3).}
    \label{fig:rate_performance_curves_cvu}
    \vspace{-2ex}
\end{figure*}
\begin{figure}[t!]
    \centering
    \subfloat[Qwen3 Feature]{\includegraphics[width=0.49\linewidth]{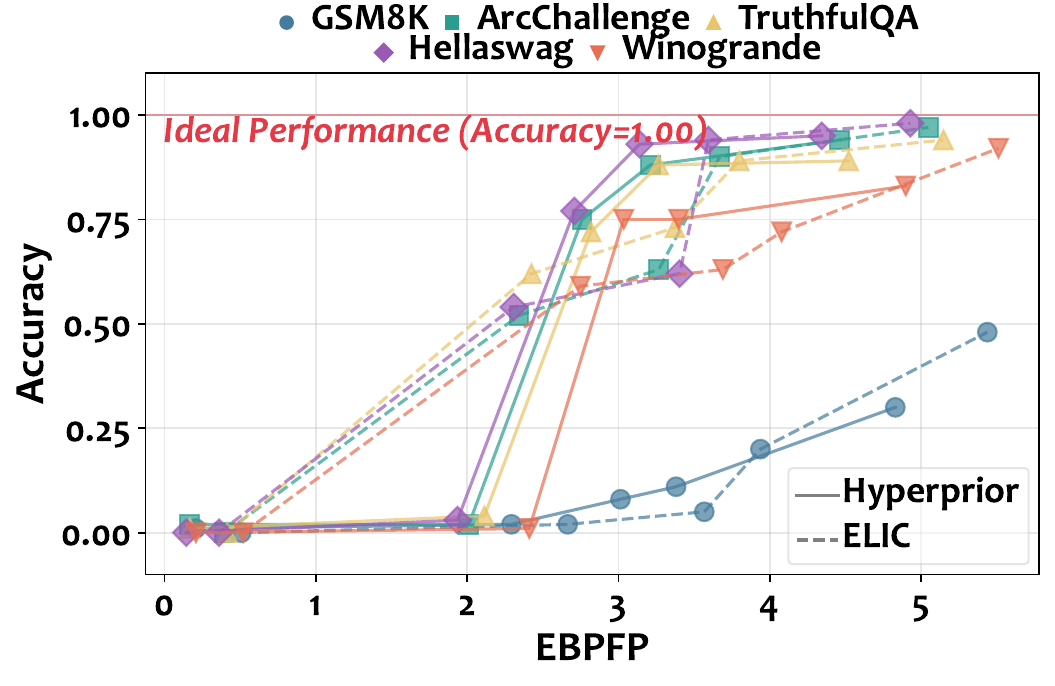}}
    \subfloat[FalconMamba Feature]{\includegraphics[width=0.49\linewidth]{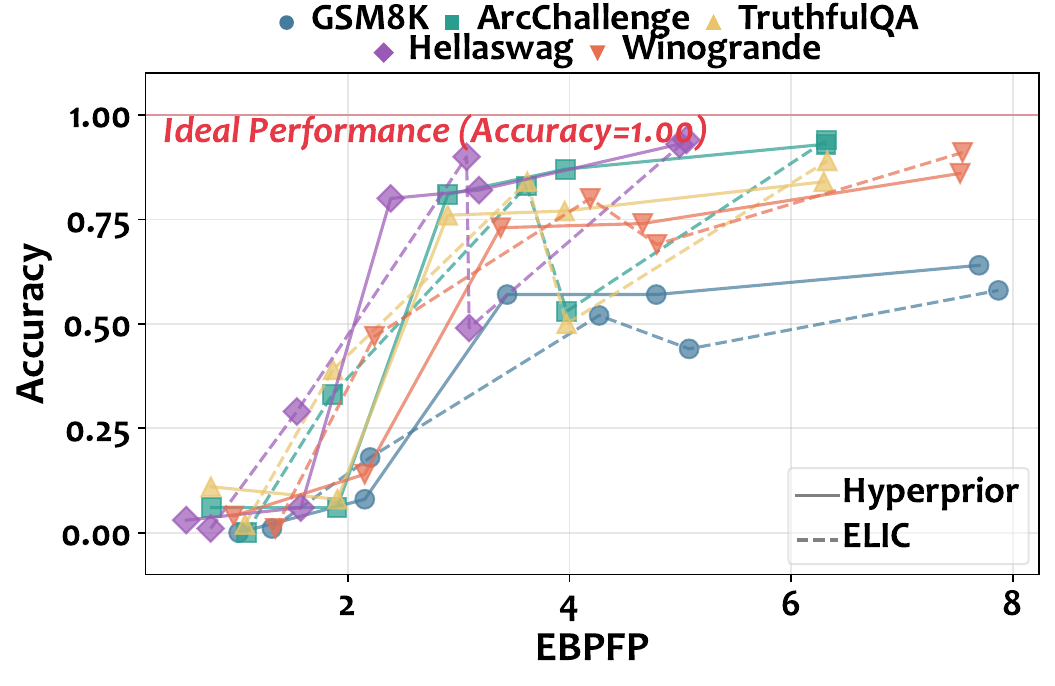}}
    \caption{Rate-performance curves for CLU (Qwen3 and FalconMamba).}
    \label{fig:rate_performance_curves_clu}
    \vspace{-2ex}
\end{figure}
\begin{figure}[t!]
    \centering
    \subfloat[ASR]{\includegraphics[width=0.49\linewidth]{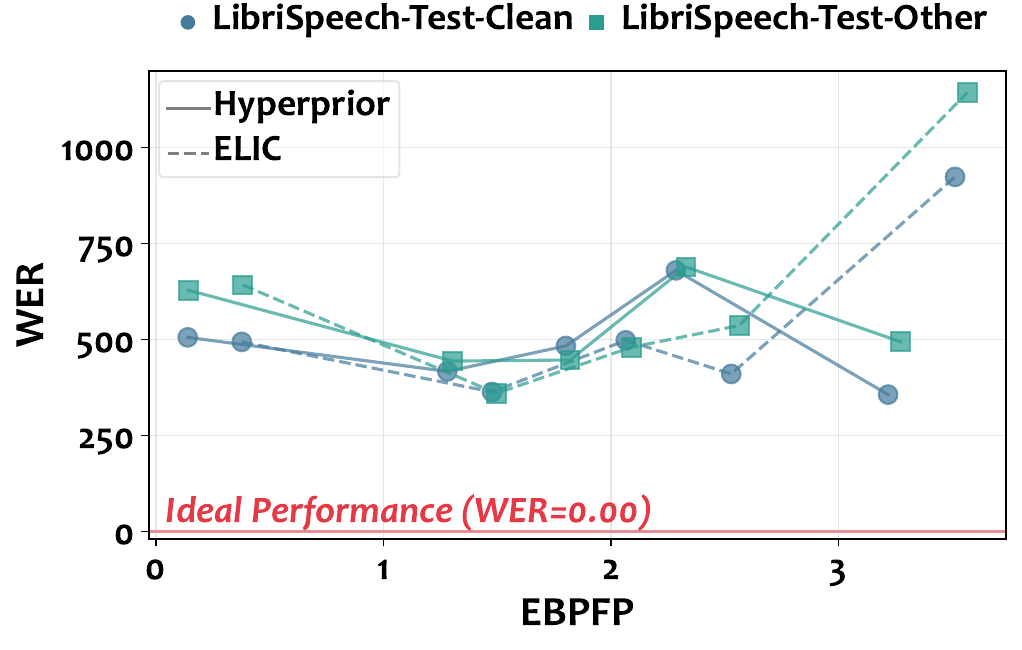}}
    \subfloat[AQA]{\includegraphics[width=0.49\linewidth]{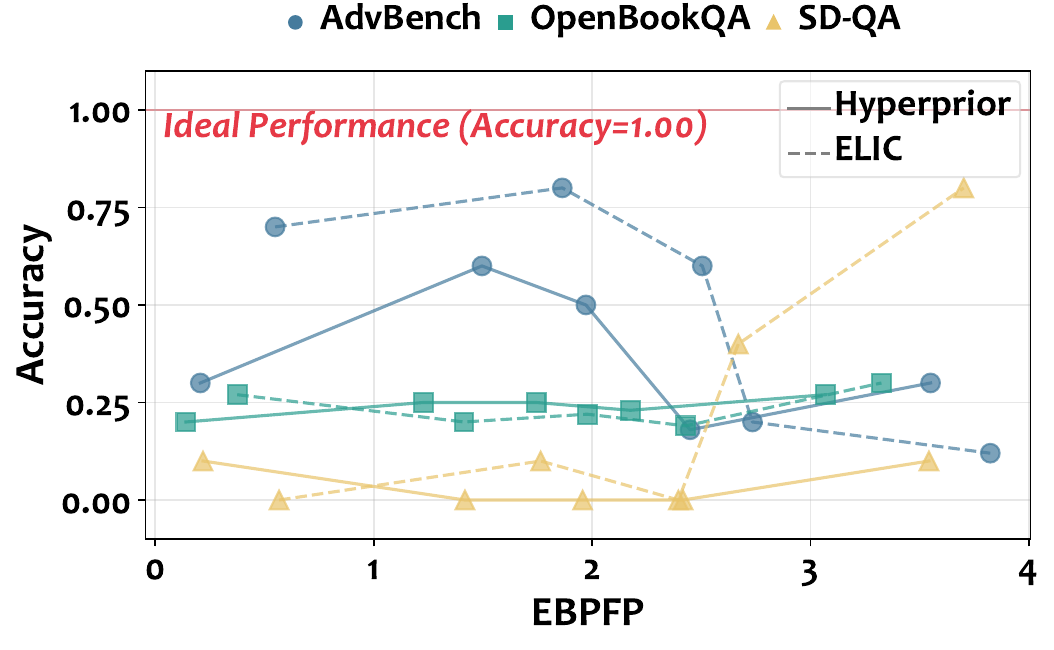}}
    \caption{Rate-performance curves for CAU (KimiAudio).}
    \label{fig:rate_performance_curves_cau}
    \vspace{-2ex}
\end{figure}
\begin{figure}[t!]
    \centering
    \subfloat[Conditioning Latents]{\includegraphics[width=0.49\linewidth]{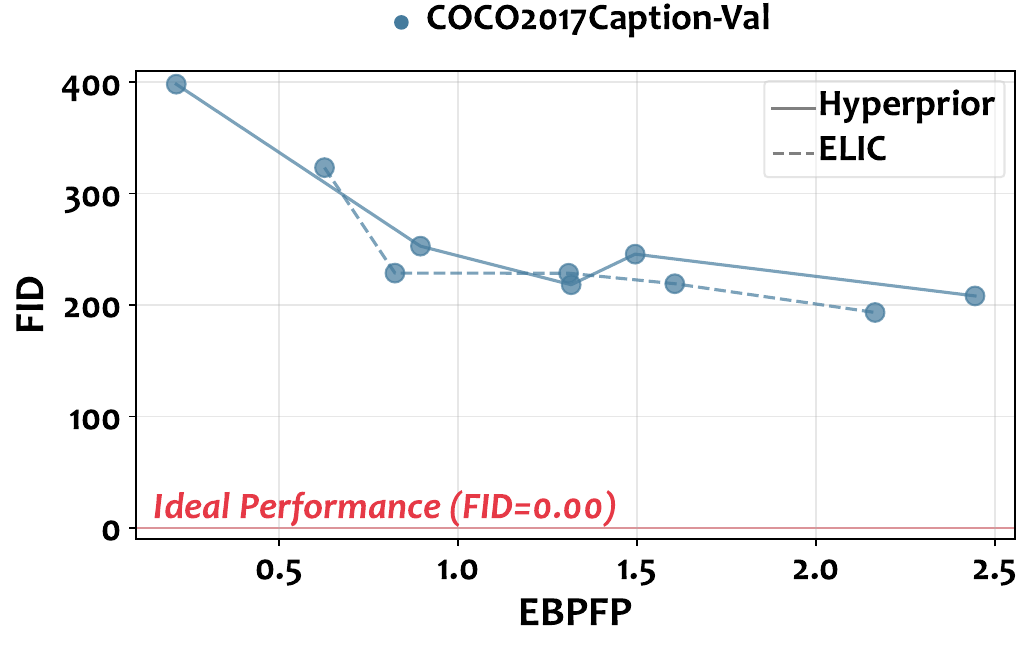}}
    \subfloat[Denoised Latents]{\includegraphics[width=0.49\linewidth]{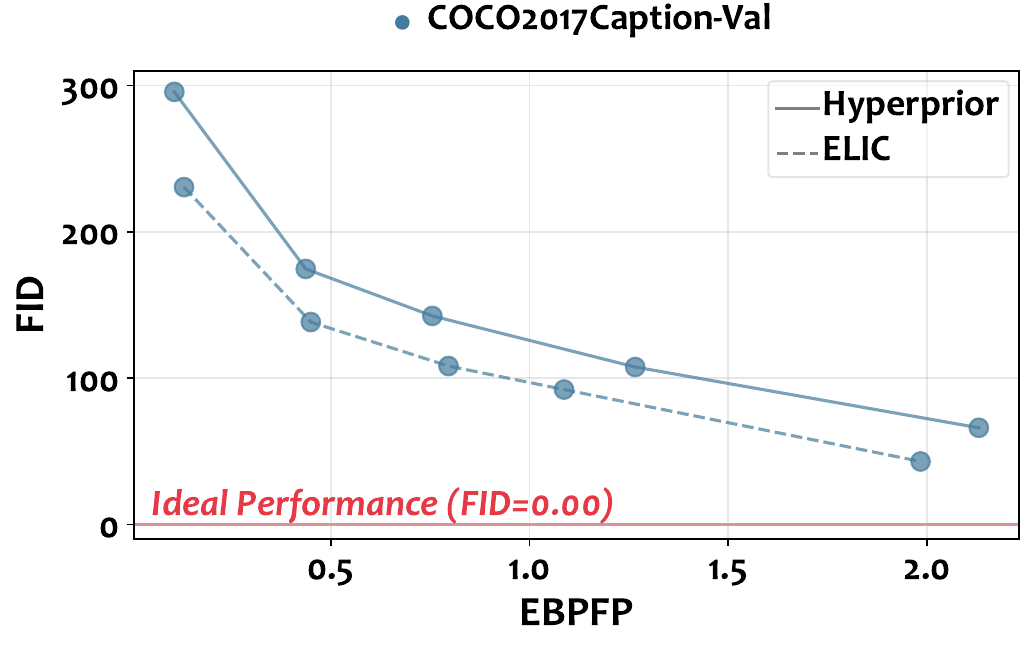}}
    \caption{Rate-performance curves for CTTI (SD3.5).}
    \label{fig:rate_performance_curves_ctti}
    \vspace{-2ex}
\end{figure}

\section{Feature-Centric Evaluation}
\label{sec:feature_centric_evaluation}

While our dataset provides broad coverage, fair benchmarking still requires standardized evaluation protocols.
To address this, we establish a targeted framework to rigorously quantify \task performance.
Our protocol shifts from traditional image-based metrics to feature-centric evaluation.

\subsection{Bitrate Measurement}
\label{sec:bitrate_measurement}

Bits per pixel (BPP) serves as the standard metric for image coding, calculated as the bitstream size divided by the number of raw pixels.
However, we argue that BPP is ill-suited for the emerging field of \task due to two primary limitations:
(1) \textit{Modality Mismatch.}
Large models process heterogeneous modality types (\eg vision, text, and audio), rendering the pixel-based definition inapplicable.
(2) \textit{Dimensionality Shift.}
BPP relies on input resolution rather than the intermediate feature dimensions.
Since features often undergo downsampling or embedding expansion, BPP fails to reflect the actual data stream density processed by the codec.
To resolve this ambiguity, we exclude BPP and propose feature-centric metrics.

\parhead{Bits Per Feature Point (BPFP).}
To address BPP's limitations, our conference version~\cite{FeatureCoding-LaMoFC} adopts BPFP as the bitrate metric.
BPFP normalizes the encoded bitstream size against the volume of the feature tensor, making it applicable across diverse modalities and architectures.
Let $N_\text{bits}$ denote the number of bits of the encoded bitstream, and $S_\text{feat}$ be the dimensions of the source feature.
BPFP is formalized as:
\begin{equation}
    \text{BPFP} = \frac{N_\text{bits}}{\prod_{d \in S_\text{feat}} d}
\end{equation}
For example, an uncompressed FP32 feature inherently has a BPFP of 32.
This metric provides a direct measure of transmission payload.

\parhead{Equivalent Bits Per Feature Point (EBPFP).}
While BPFP measures absolute bandwidth, it is insufficient for evaluating algorithmic efficiency when input precisions vary.
Our dataset comprises features with different precisions, including full-precision (FP32) and half-precision (FP16/BF16).
Directly comparing BPFP across these precisions is biased, as it obscures that achieving a BPFP of 2.0 on a 16-bit source (8:1 compression) implies lower coding efficiency than on a 32-bit source (16:1 compression).
To decouple bitrate savings derived from source quantization versus actual codec optimization, we introduce EBPFP.
It projects the bitrate onto a standard 32-bit equivalent scale.
Let $P_\text{raw}$ denote the bit depth of the raw feature elements (\eg 32 or 16).
We first define the \emph{equivalent bits} $N_\text{eq} = N_\text{bits} \times \frac{32}{P_\text{raw}}$, which scales the actual bitstream size to a 32-bit baseline.
Consequently, EBPFP is formulated as the equivalent bits per feature point:
\begin{equation}
    \text{EBPFP}
    = \frac{N_\text{eq}}{\prod_{d \in S_\text{feat}} d}
    = \frac{32}{P_\text{raw}} \frac{N_\text{bits}}{\prod_{d \in S_\text{feat}} d}
    = \frac{32}{P_\text{raw}} \text{BPFP}
    \label{equ:ebpfp}
\end{equation}
This metric effectively normalizes the compression ratio to a common 32-bit baseline, allowing for rigorous assessment of feature coding efficiency independent of the raw precision.

\begin{table*}[t!]
    \centering
    \caption{Rate-performance evaluation for CVU (DINOv3).
        Pearson correlation coefficient $\rho$ is used to measure the correlation between feature reconstruction error (MSE) and downstream task performance.}
    \label{tab:dinov3_rate_accuracy}
    \resizebox{\linewidth}{!}{

\begin{tabular}{ lc|rrrr|rrrr|rrrr|rrrr|rrrr }
    \toprule
                                         &
                                         &
    \multicolumn{4}{c|}{ImageNet-Val}    &
    \multicolumn{4}{c|}{ImageNet-A}      &
    \multicolumn{4}{c|}{ImageNet-R}      &
    \multicolumn{4}{c|}{ADE20K-Val}      &
    \multicolumn{4}{c}{NYUDepthV2-Test}                                                                                     \\
                                         & $\lambda$
                                         & EBPFP$\downarrow$ & $\mathcal{A}\uparrow$ & MSE$\downarrow$ & B$_{\max}\uparrow$
                                         & EBPFP$\downarrow$ & $\mathcal{A}\uparrow$ & MSE$\downarrow$ & B$_{\max}\uparrow$
                                         & EBPFP$\downarrow$ & $\mathcal{A}\uparrow$ & MSE$\downarrow$ & B$_{\max}\uparrow$
                                         & EBPFP$\downarrow$ & mIoU$\uparrow$        & MSE$\downarrow$ & B$_{\max}\uparrow$
                                         & EBPFP$\downarrow$ & RMSE$\downarrow$      & MSE$\downarrow$ & B$_{\max}\uparrow$ \\
    \midrule
    Original                             &
                                         & 32                & 1.000                 & 0.000           & $\infty$
                                         & 32                & 1.000                 & 0.000           & $\infty$
                                         & 32                & 1.000                 & 0.000           & $\infty$
                                         & 32                & 0.782                 & 0.000           & $\infty$
                                         & 32                & 0.132                 & 0.000           & $\infty$           \\
    \midrule
    \multirow{5}{*}{Hyperprior}          & 0.001
                                         & 0.039             & 0.000                 & 14.867          & 171.934
                                         & 0.038             & 0.010                 & 14.832          & 174.148
                                         & 0.039             & 0.000                 & 14.837          & 172.675
                                         & 0.052             & 0.332                 & 4861.017        & 166.114
                                         & 0.052             & 0.271                 & 4767.951        & 161.694            \\

                                         & 0.004
                                         & 0.224             & 0.050                 & 20.268          & 136.428
                                         & 0.218             & 0.010                 & 16.535          & 171.317
                                         & 0.223             & 0.070                 & 17.839          & 171.599
                                         & 0.371             & 0.673                 & 716.643         & 161.760
                                         & 0.372             & 0.154                 & 728.699         & 165.168            \\

                                         & 0.007
                                         & 0.423             & 0.590                 & 158.940         & 166.902
                                         & 0.417             & 0.290                 & 124.943         & 167.018
                                         & 0.417             & 0.720                 & 156.594         & 165.607
                                         & 0.671             & 0.747                 & 467.635         & 159.960
                                         & 0.672             & 0.135                 & 417.689         & 163.991            \\

                                         & 0.010
                                         & 0.754             & 0.690                 & 402.420         & 164.595
                                         & 0.752             & 0.690                 & 362.934         & 165.545
                                         & 0.741             & 0.920                 & 415.583         & 162.709
                                         & 0.878             & 0.755                 & 601.756         & 158.734
                                         & 0.879             & 0.132                 & 522.778         & 153.150            \\

                                         & 0.020
                                         & 1.289             & 0.910                 & 172.198         & 157.984
                                         & 1.296             & 0.940                 & 139.992         & 157.457
                                         & 1.270             & 0.980                 & 156.691         & 158.490
                                         & 1.306             & 0.772                 & 377.352         & 149.704
                                         & 1.306             & 0.131                 & 335.304         & 158.854            \\

    \rowcolor{gray!20}
    \multicolumn{2}{l|}{Correlation}     &
    \multicolumn{4}{r|}{$\rho$ = +0.727} &
    \multicolumn{4}{r|}{$\rho$ = +0.690} &
    \multicolumn{4}{r|}{$\rho$ = +0.784} &
    \multicolumn{4}{r|}{$\rho$ = -0.988} &
    \multicolumn{4}{r}{$\rho$ = +0.996}                                                                                     \\

    \midrule
    \multirow{5}{*}{ELIC}                & 0.001
                                         & 0.027             & 0.010                 & 13.686          & 53.336
                                         & 0.025             & 0.010                 & 13.612          & 53.442
                                         & 0.027             & 0.000                 & 13.679          & 53.496
                                         & 0.059             & 0.324                 & 4037.896        & 50.323
                                         & 0.056             & 0.279                 & 4074.353        & 50.513             \\

                                         & 0.004
                                         & 0.130             & 0.190                 & 12.266          & 52.439
                                         & 0.120             & 0.080                 & 12.510          & 52.397
                                         & 0.129             & 0.260                 & 12.409          & 52.775
                                         & 0.343             & 0.678                 & 545.571         & 50.082
                                         & 0.333             & 0.150                 & 539.597         & 49.953             \\

                                         & 0.007
                                         & 0.361             & 0.770                 & 37.192          & 52.184
                                         & 0.345             & 0.420                 & 27.417          & 52.218
                                         & 0.357             & 0.820                 & 28.220          & 51.827
                                         & 0.545             & 0.707                 & 659.724         & 49.441
                                         & 0.534             & 0.140                 & 610.670         & 49.185             \\

                                         & 0.010
                                         & 0.570             & 0.900                 & 43.415          & 51.348
                                         & 0.559             & 0.820                 & 39.739          & 51.721
                                         & 0.559             & 0.990                 & 45.221          & 51.689
                                         & 0.783             & 0.727                 & 743.415         & 49.218
                                         & 0.753             & 0.137                 & 702.785         & 49.014             \\

                                         & 0.020
                                         & 1.210             & 1.000                 & 11.530          & 49.882
                                         & 1.213             & 1.000                 & 8.637           & 49.901
                                         & 1.190             & 1.000                 & 9.191           & 50.060
                                         & 1.305             & 0.771                 & 295.001         & 48.138
                                         & 1.304             & 0.132                 & 304.865         & 47.877             \\
    \rowcolor{gray!20}
    \multicolumn{2}{l|}{Correlation}     &
    \multicolumn{4}{r|}{$\rho$ = +0.508} &
    \multicolumn{4}{r|}{$\rho$ = +0.281} &
    \multicolumn{4}{r|}{$\rho$ = +0.494} &
    \multicolumn{4}{r|}{$\rho$ = -0.988} &
    \multicolumn{4}{r}{$\rho$ = +0.994}                                                                                     \\
    \bottomrule
\end{tabular}
}
    \vspace{-2ex}
\end{table*}

\subsection{Feature Coding Pipeline}
\label{sec:feature_coding_pipeline}

The pipeline generally comprises three sequential stages, \ie pre-processing, codec, and post-processing.
In pre-processing, the original features are quantized to integer values represented with a specified bit-depth (\eg 8 bits~\cite{FeatureCoding-LaMoFC}).%
\footnote{We omit the outlier truncation typically used prior to quantization, as it requires complex manual tuning and high experimental costs, favoring a flexible non-linear transform strategy~\cite{FeatureCoding-DT-UFC} instead.}
Subsequently, the quantized data is packed to align with the standard input format of the codec.
The codec comprises two stages: encoding and decoding.
It takes the packed quantized feature as input, first compressing it into an encoded bitstream and subsequently reconstructing it into the decoded feature.
Finally, in post-processing, the operation is reversed: the decoded feature is unpacked to recover the quantized sequence, which is then de-quantized to produce the final feature with the original numeric precision.

\parhead{Packing Details.}
To accommodate the diverse feature structures inherent in different large models, the packing mechanism standardizes heterogeneous tensors into a unified format compatible with the core codec.
Features ranging from multi-dimensional key/value caches to 1D sentence embeddings are first reshaped into a unified 2D layout (typically $N \times C$ as in~\cite{FeatureCoding-LaMoFC}).
This transformation maps high-dimensional feature data onto a 2D grid, treating sequence length ($N$) and channel dimension ($C$) as conceptually spatial height and width, thereby aligning with the input standard of codecs.
In this paper, we use a straightforward per-tensor packing strategy, which treats each feature tensor independently.
It preserves the modularity of the original feature structure and allows for granular access.
The auxiliary metadata, such as original shapes and grouping indices, can be recorded during the initialization phase to ensure accurately reversible unpacking for subsequent coding.
Therefore, the metadata transmission overhead can be considered negligible during the evaluation.

\parhead{Coding Settings.}
We prioritize neural codecs over traditional hand-crafted codecs.
The latter are often optimized for CPU-centric workflows and cannot fully exploit the massive parallelism of GPUs, making them inefficient for modern AI-native transmission pipelines.
Thus, we select representative learning-based image codecs, specifically Hyperprior~\cite{FeatureCoding-Hyperprior} and ELIC~\cite{FeatureCoding-ELIC}, as our baselines.%
\footnote{Since \task is an emerging field lacking established native feature codecs (as discussed in~\cref{sec:towards_native_feature_coding}), adapting existing image-centric schemes serves as a reasonable starting point as done in~\cite{FeatureCoding-LaMoFC,FeatureCoding-DT-UFC}.}
The input and output layers of these models are modified to accept and output a single channel to match the 2D packed feature format.
These methods are implemented within the universal feature transformation framework~\cite{FeatureCoding-DT-UFC} to avoid complexity in pre-/post-processing.
For both codecs, the rate-controlling parameter $\lambda$ is standardized to $\{0.001, 0.004, 0.007, 0.01, 0.02\}$, and we employ the official pre-trained weights~\cite{FeatureCoding-DT-UFC} without fine-tuning.

\subsection{Experimental Setup}
\label{sec:experimental_setup}

\parhead{Dataset.}
Our benchmark is established to evaluate \task performance, utilizing intermediate features extracted from several representative large models.
Since distribution alignment is required~\cite{FeatureCoding-DT-UFC}, we additionally provide an auxiliary calibration set alongside the standard test split.
Specifically, for each model, we collect ten samples randomly selected from non-test splits of source data.
Their features are used solely to derive the data-driven non-uniform transformation~\cite{FeatureCoding-DT-UFC}.
For rigorous evaluation, the transformation is constructed solely from the auxiliary set and then frozen as a fixed pre-processing step during benchmark testing.

\parhead{Metrics.}
We comprehensively assess \textit{efficiency}, \textit{distortion}, \textit{efficacy}, and \textit{practicality} of the coding scheme across all scenarios.
For coding efficiency, we employ EBPFP defined in~\cref{equ:ebpfp}, which normalizes the bitstream length by the number of feature elements and the raw precision to enable resolution-agnostic comparisons.
MSE is utilized to quantify the element-wise reconstruction distortion relative to the original uncompressed features.
Since the primary objective is to preserve downstream capabilities, we report the task-specific performance metrics for each model, as listed in~\cref{tab:table_summary}.
We evaluate overall efficacy via the rate-performance relationship obtained by varying the bitrate constraint (\ie the rate-controlling parameter $\lambda$), enabling direct comparison of task performance at similar compression levels.
In addition to these standard metrics, we assess codec practicality using the maximum advantageous bandwidth ($\text{B}_{\max}$ as stated in~\cref{sec:practicality_analysis}) and peak GPU memory during runtime.
Rather than reporting isolated inference latency, we adopt $\text{B}_{\max}$ to reflect practical constraints by accounting for the end-to-end time of codec encoding, bitstream transmission, and decoding.

\parhead{Implementation.}
To comprehensively evaluate the \task performance of the pretrained codec baselines~\cite{FeatureCoding-Hyperprior,FeatureCoding-ELIC}, we conduct evaluations on an NVIDIA RTX 4090 GPU with a batch size of 1.
This configuration ensures that the peak memory footprints primarily reflect the pure memory requirements of the respective codecs.
To guarantee deterministic and reproducible outputs, we employ greedy decoding (\eg setting the temperature to 0) for Qwen3, FalconMamba, and KimiAudio.
For DINOv3 and SD3.5, we strictly adhere to the resolution and preprocessing settings recommended in the original papers without any additional adjustments, thereby accurately reflecting their intrinsic performance.

\begin{table*}[t!]
    \centering
    \caption{Rate-performance evaluation for CLU (Qwen3).}
    \label{tab:qwen_rate_accuracy}
    \resizebox{\linewidth}{!}{
\begin{tabular}{ lc|rrrr|rrrr|rrrr|rrrr|rrrr }
    \toprule
                                         &                   &
    \multicolumn{4}{c}{GSM8K}            &
    \multicolumn{4}{c}{ArcChallenge}     &
    \multicolumn{4}{c}{TruthfulQA}       &
    \multicolumn{4}{c}{Hellaswag}        &
    \multicolumn{4}{c}{Winogrande}                                                                                          \\
                                         & $\lambda$
                                         & EBPFP$\downarrow$ & $\mathcal{A}\uparrow$ & MSE$\downarrow$ & B$_{\max}\uparrow$
                                         & EBPFP$\downarrow$ & $\mathcal{A}\uparrow$ & MSE$\downarrow$ & B$_{\max}\uparrow$
                                         & EBPFP$\downarrow$ & $\mathcal{A}\uparrow$ & MSE$\downarrow$ & B$_{\max}\uparrow$
                                         & EBPFP$\downarrow$ & $\mathcal{A}\uparrow$ & MSE$\downarrow$ & B$_{\max}\uparrow$
                                         & EBPFP$\downarrow$ & $\mathcal{A}\uparrow$ & MSE$\downarrow$ & B$_{\max}\uparrow$ \\
    \midrule
    Original                             &
                                         & 32                & 1.000                 & 0.000           & $\infty$
                                         & 32                & 1.000                 & 0.000           & $\infty$
                                         & 32                & 1.000                 & 0.000           & $\infty$
                                         & 32                & 1.000                 & 0.000           & $\infty$
                                         & 32                & 1.000                 & 0.000           & $\infty$           \\
    \midrule
    \multirow{5}{*}{Hyperprior}          & 0.001
                                         & 0.206             & 0.010                 & 9.210           & 45.511
                                         & 0.166             & 0.020                 & 6.646           & 62.907
                                         & 0.169             & 0.010                 & 6.745           & 61.911
                                         & 0.144             & 0.000                 & 5.091           & 72.820
                                         & 0.208             & 0.000                 & 9.537           & 43.809             \\
                                         & 0.004
                                         & 2.293             & 0.020                 & 2.249           & 41.365
                                         & 2.012             & 0.020                 & 1.924           & 58.447
                                         & 2.115             & 0.040                 & 1.969           & 57.293
                                         & 1.936             & 0.030                 & 1.675           & 68.036
                                         & 2.411             & 0.010                 & 2.398           & 39.613             \\
                                         & 0.007
                                         & 3.014             & 0.080                 & 1.263           & 40.141
                                         & 2.760             & 0.750                 & 1.075           & 55.669
                                         & 2.819             & 0.720                 & 1.083           & 54.666
                                         & 2.709             & 0.770                 & 0.963           & 65.006
                                         & 3.036             & 0.750                 & 1.350           & 38.559             \\
                                         & 0.010
                                         & 3.381             & 0.110                 & 0.863           & 39.269
                                         & 3.210             & 0.880                 & 0.748           & 54.460
                                         & 3.266             & 0.880                 & 0.750           & 53.431
                                         & 3.140             & 0.930                 & 0.669           & 62.861
                                         & 3.399             & 0.750                 & 0.878           & 37.759             \\
                                         & 0.020
                                         & 4.830             & 0.300                 & 0.512           & 17.581
                                         & 4.462             & 0.940                 & 0.459           & 51.331
                                         & 4.522             & 0.890                 & 0.458           & 15.232
                                         & 4.344             & 0.950                 & 0.436           & 25.606
                                         & 4.899             & 0.830                 & 0.513           & 11.007             \\
    \rowcolor{gray!20}
    \multicolumn{2}{l|}{Correlation}     &
    \multicolumn{4}{r|}{$\rho$ = -0.575} &
    \multicolumn{4}{r|}{$\rho$ = -0.758} &
    \multicolumn{4}{r|}{$\rho$ = -0.774} &
    \multicolumn{4}{r|}{$\rho$ = -0.789} &
    \multicolumn{4}{r}{$\rho$ = -0.745}                                                                                     \\
    \midrule
    \multirow{5}{*}{ELIC}                & 0.001
                                         & 0.508             & 0.000                 & 3.267           & 7.647
                                         & 0.399             & 0.000                 & 2.725           & 12.458
                                         & 0.415             & 0.000                 & 2.765           & 12.307
                                         & 0.361             & 0.000                 & 2.666           & 17.765
                                         & 0.520             & 0.000                 & 3.433           & 7.266              \\
                                         & 0.004
                                         & 2.665             & 0.020                 & 1.070           & 7.047
                                         & 2.344             & 0.520                 & 1.011           & 11.343
                                         & 2.426             & 0.620                 & 1.020           & 11.265
                                         & 2.308             & 0.540                 & 0.985           & 16.372
                                         & 2.749             & 0.590                 & 1.055           & 6.701              \\
                                         & 0.007
                                         & 3.568             & 0.050                 & 0.960           & 6.784
                                         & 3.267             & 0.630                 & 0.890           & 10.920
                                         & 3.374             & 0.730                 & 0.902           & 10.810
                                         & 3.405             & 0.620                 & 0.850           & 15.631
                                         & 3.691             & 0.630                 & 0.941           & 6.353              \\
                                         & 0.010
                                         & 3.939             & 0.200                 & 0.769           & 6.590
                                         & 3.670             & 0.900                 & 0.696           & 10.723
                                         & 3.799             & 0.890                 & 0.703           & 10.495
                                         & 3.595             & 0.940                 & 0.674           & 15.423
                                         & 4.079             & 0.720                 & 0.751           & 6.259              \\
                                         & 0.020
                                         & 5.439             & 0.480                 & 0.438           & 6.222
                                         & 5.049             & 0.970                 & 0.415           & 10.137
                                         & 5.148             & 0.940                 & 0.411           & 10.000
                                         & 4.930             & 0.980                 & 0.474           & 14.602
                                         & 5.512             & 0.920                 & 0.412           & 5.924              \\
    \rowcolor{gray!20}
    \multicolumn{2}{l|}{Correlation}     &
    \multicolumn{4}{r|}{$\rho$ = -0.601} &
    \multicolumn{4}{r|}{$\rho$ = -0.962} &
    \multicolumn{4}{r|}{$\rho$ = -0.991} &
    \multicolumn{4}{r|}{$\rho$ = -0.953} &
    \multicolumn{4}{r}{$\rho$ = -0.984}                                                                                     \\
    \bottomrule
\end{tabular}

}
    \vspace{-2ex}
\end{table*}
\begin{table*}[t!]
    \centering
    \caption{Rate-performance evaluation for CLU (FalconMamba).}
    \label{tab:falconmamba_rate_accuracy}
    \resizebox{\linewidth}{!}{

\begin{tabular}{ lc|rrrr|rrrr|rrrr|rrrr|rrrr }
    \toprule
                                         &
                                         &
    \multicolumn{4}{c}{GSM8K}            &
    \multicolumn{4}{c}{ArcChallenge}     &
    \multicolumn{4}{c}{TruthfulQA}       &
    \multicolumn{4}{c}{Hellaswag}        &
    \multicolumn{4}{c}{Winogrande}                                                                                          \\
                                         & $\lambda$
                                         & EBPFP$\downarrow$ & $\mathcal{A}\uparrow$ & MSE$\downarrow$ & B$_{\max}\uparrow$
                                         & EBPFP$\downarrow$ & $\mathcal{A}\uparrow$ & MSE$\downarrow$ & B$_{\max}\uparrow$
                                         & EBPFP$\downarrow$ & $\mathcal{A}\uparrow$ & MSE$\downarrow$ & B$_{\max}\uparrow$
                                         & EBPFP$\downarrow$ & $\mathcal{A}\uparrow$ & MSE$\downarrow$ & B$_{\max}\uparrow$
                                         & EBPFP$\downarrow$ & $\mathcal{A}\uparrow$ & MSE$\downarrow$ & B$_{\max}\uparrow$ \\
    \midrule
    Original                             &
                                         & 32                & 1.000                 & 0.000           & $\infty$
                                         & 32                & 1.000                 & 0.000           & $\infty$
                                         & 32                & 1.000                 & 0.000           & $\infty$
                                         & 32                & 1.000                 & 0.000           & $\infty$
                                         & 32                & 1.000                 & 0.000           & $\infty$           \\
    \midrule
    \multirow{5}{*}{Hyperprior}          & 0.001
                                         & 1.015             & 0.000                 & 0.116           & 20.358  
                                         & 0.770             & 0.060                 & 0.072           & 29.520  
                                         & 0.765             & 0.110                 & 0.072           & 29.866  
                                         & 0.543             & 0.030                 & 0.050           & 43.254  
                                         & 0.971             & 0.040                 & 0.091           & 21.823       \\
                                         & 0.004
                                         & 2.153             & 0.080                 & 0.089           & 19.405  
                                         & 1.903             & 0.060                 & 0.052           & 28.231  
                                         & 1.907             & 0.080                 & 0.051           & 28.172  
                                         & 1.576             & 0.060                 & 0.036           & 41.519  
                                         & 2.153             & 0.140                 & 0.065           & 20.827       \\
                                         & 0.007
                                         & 3.437             & 0.570                 & 0.079           & 18.633  
                                         & 2.903             & 0.810                 & 0.042           & 27.204  
                                         & 2.901             & 0.760                 & 0.041           & 27.538  
                                         & 2.387             & 0.800                 & 0.028           & 40.003  
                                         & 3.379             & 0.730                 & 0.052           & 19.828       \\
                                         & 0.010
                                         & 4.783             & 0.570                 & 0.062           & 17.792  
                                         & 3.962             & 0.870                 & 0.032           & 26.379  
                                         & 3.956             & 0.770                 & 0.031           & 26.652  
                                         & 3.183             & 0.820                 & 0.022           & 38.949  
                                         & 4.661             & 0.740                 & 0.040           & 19.168        \\
                                         & 0.020
                                         & 7.694             & 0.640                 & 0.041           & 16.028  
                                         & 6.314             & 0.930                 & 0.017           & 24.454  
                                         & 6.293             & 0.840                 & 0.017           & 24.569  
                                         & 4.991             & 0.930                 & 0.012           & 36.321  
                                         & 7.523             & 0.860                 & 0.022           & 17.278       \\
    \rowcolor{gray!20}
    \multicolumn{2}{l|}{Correlation}     &
    \multicolumn{4}{r|}{$\rho$ = -0.874} &
    \multicolumn{4}{r|}{$\rho$ = -0.876} &
    \multicolumn{4}{r|}{$\rho$ = -0.859} &
    \multicolumn{4}{r|}{$\rho$ = -0.888} &
    \multicolumn{4}{r}{$\rho$ = -0.914}                                                                                     \\
    \midrule
    \multirow{5}{*}{ELIC}                & 0.001
                                         & 1.316             & 0.010                 & 0.100           & 4.637
                                         & 1.084             & 0.000                 & 0.061           & 6.948
                                         & 1.073             & 0.020                 & 0.061           & 6.784
                                         & 0.763             & 0.010                 & 0.042           & 10.280
                                         & 1.346             & 0.010                 & 0.077           & 5.012              \\
                                         & 0.004
                                         & 2.201             & 0.180                 & 0.076           & 4.814
                                         & 1.859             & 0.330                 & 0.038           & 7.152
                                         & 1.867             & 0.390                 & 0.038           & 6.954
                                         & 1.541             & 0.290                 & 0.026           & 10.763
                                         & 2.241             & 0.470                 & 0.048           & 4.926              \\
                                         & 0.007
                                         & 5.079             & 0.440                 & 0.066           & 4.436
                                         & 3.975             & 0.530                 & 0.030           & 6.721
                                         & 3.974             & 0.500                 & 0.030           & 6.840
                                         & 3.095             & 0.490                 & 0.021           & 10.311
                                         & 4.788             & 0.690                 & 0.038           & 4.773              \\
                                         & 0.010
                                         & 4.267             & 0.520                 & 0.054           & 4.489
                                         & 3.614             & 0.830                 & 0.023           & 6.743
                                         & 3.619             & 0.840                 & 0.023           & 6.800
                                         & 3.072             & 0.900                 & 0.016           & 10.225
                                         & 4.187             & 0.800                 & 0.029           & 4.798              \\
                                         & 0.020
                                         & 7.869             & 0.580                 & 0.025           & 4.022
                                         & 6.318             & 0.940                 & 0.010           & 6.262
                                         & 6.325             & 0.890                 & 0.010           & 6.340
                                         & 5.055             & 0.940                 & 0.007           & 9.733
                                         & 7.545             & 0.910                 & 0.013           & 4.328              \\
    \rowcolor{gray!20}
    \multicolumn{2}{l|}{Correlation}     &
    \multicolumn{4}{r|}{$\rho$ = -0.924} &
    \multicolumn{4}{r|}{$\rho$ = -0.980} &
    \multicolumn{4}{r|}{$\rho$ = -0.974} &
    \multicolumn{4}{r|}{$\rho$ = -0.954} &
    \multicolumn{4}{r}{$\rho$ = -0.985}                                                                                     \\
    \bottomrule
\end{tabular}
}
    \vspace{-2ex}
\end{table*}

\subsection{Results and Discussion}
\label{sec:results_and_discussion}

\subsubsection{Rate-Performance (R-P) Analysis}
\label{sec:rate_accuracy}

The R-P evaluation reveals distinct patterns across different modalities and architectures (see~\cref{fig:rate_performance_curves_cvu,fig:rate_performance_curves_clu,fig:rate_performance_curves_cau,fig:rate_performance_curves_ctti} for curves and~\cref{tab:dinov3_rate_accuracy,tab:qwen_rate_accuracy,tab:falconmamba_rate_accuracy,tab:kimiaudio_rate_accuracy,tab:sd35cond_rate_accuracy} for detailed values).
In both CVU and CLU tasks, the ELIC baseline consistently matches or outperforms the Hyperprior counterpart in downstream performance.
At a macro level, visual features exhibit relatively strong adaptability to feature coding, closely approximating their original performance at larger $\lambda$ values.
However, in language understanding, the overall performance recovery is comparatively limited, reflecting the greater sensitivity of language features.
Significant divergences emerge in multi-modal and generative settings.
Specifically, in audio understanding (\cref{tab:kimiaudio_rate_accuracy}), applying either codec triggers severe functional degradation, causing the model to lose its original capabilities.
This suggests that internal multi-modal mixed features are highly sensitive to the current feature coding pipeline, which disrupts the delicate representations required for audio reasoning.
Furthermore, the CTTI results shown in~\cref{tab:sd35cond_rate_accuracy}, underscore the critical role of feature positioning.
Compressing conditioning representations (including both text embeddings and image latents) leads to substantial distribution shifts.
Since these representations serve as precise guidance for the denoising process, they possess a very low tolerance for perturbation.
In contrast, the denoised latents exhibit a high degree of inherent redundancy, as reflected in redundancy analyses (\cref{sec:redundancy_analysis,tab:sd35cond_feature_analysis}).
This redundancy provides a solid foundation for feature coding, enabling codecs to effectively preserve the generative distribution.

\subsubsection{Distortion-Performance (D-P) Analysis}
\label{sec:distortion_performance_analysis}

We quantitatively evaluate the alignment between feature reconstruction error (MSE) and downstream task performance, measured by the Pearson correlation coefficient $\rho$, detailed in the ``Correlation'' rows of \cref{tab:dinov3_rate_accuracy,tab:qwen_rate_accuracy,tab:falconmamba_rate_accuracy,tab:kimiaudio_rate_accuracy,tab:sd35cond_rate_accuracy}.
For spatially structured representations, such as depth estimation and semantic segmentation ($|\rho| > 0.95$) in~\cref{tab:dinov3_rate_accuracy}, as well as most language tasks ($|\rho| > 0.7$) in~\cref{tab:qwen_rate_accuracy,tab:falconmamba_rate_accuracy}, we observe strong statistical alignment.
This suggests that for these features, element-wise MSE serves as a reliable proxy for downstream task utility.
However, a notable divergence emerges in classification tasks when using shallow features (\ie Layer 10 in~\cref{tab:dinov3_rate_accuracy}).
Specifically, the correlation weakens substantially under the ELIC codec ($\rho$ drops to $0.28$).
This reveals that MSE is a poor proxy for downstream performance on shallow layers.
Since shallow features encode a complex mixture of dense background noise and low-level details, the element-wise MSE indiscriminately penalizes all distortions.
It struggles to differentiate between benign variations in task-irrelevant noise and critical degradations of foundational structural elements.
Furthermore, we observe highly unstable correlations in~\cref{tab:kimiaudio_rate_accuracy}, where downstream capability experiences a collapse under severe compression.
In~\cref{tab:sd35cond_rate_accuracy}, multi-modal conditioning signals also exhibit inconsistent alignment ($\rho=0.874$ for Hyperprior \vs $0.216$ for ELIC), indicating a severe decoupling between MSE and semantic integrity.
In contrast, the denoised latents maintain an almost perfect correlation ($\rho \approx 1$).
These results highlight a critical limitation in current distortion evaluation paradigms.
While MSE effectively tracks gradual degradation in deep abstract semantics and redundant task-specific dense predictions (\eg depth estimation and segmentation maps), it falls short and severely decouples from downstream utility for complex multi-modal alignments or shallow visual representations.
Therefore, developing new, semantics-aware distortion metrics is imperative, particularly to accurately perceive the tipping point where feature coding triggers a catastrophic collapse in model capability.

\begin{table*}[t!]
    \centering
    \caption{Rate-performance evaluation for CAU (KimiAudio).}
    \label{tab:kimiaudio_rate_accuracy}
    \resizebox{\linewidth}{!}{
\begin{tabular}{ lc|rrrr|rrrr|rrrr|rrrr|rrrr }
    \toprule
                                                &
                                                &
    \multicolumn{4}{c|}{LibriSpeech-Test-Clean} &
    \multicolumn{4}{c|}{LibriSpeech-Test-Other} &
    \multicolumn{4}{c|}{AdvBench}               &
    \multicolumn{4}{c|}{OpenBookQA}             &
    \multicolumn{4}{c}{SD-QA}                                                                                                      \\
                                                & $\lambda$
                                                & EBPFP$\downarrow$ & WER$\downarrow$       & MSE$\downarrow$ & B$_{\max}\uparrow$
                                                & EBPFP$\downarrow$ & WER$\downarrow$       & MSE$\downarrow$ & B$_{\max}\uparrow$
                                                & EBPFP$\downarrow$ & $\mathcal{A}\uparrow$ & MSE$\downarrow$ & B$_{\max}\uparrow$
                                                & EBPFP$\downarrow$ & $\mathcal{A}\uparrow$ & MSE$\downarrow$ & B$_{\max}\uparrow$
                                                & EBPFP$\downarrow$ & $\mathcal{A}\uparrow$ & MSE$\downarrow$ & B$_{\max}\uparrow$ \\
    \midrule
    Original                                    &
                                                & 32                & 0.000                 & 0.000           & $\infty$
                                                & 32                & 0.000                 & 0.000           & $\infty$
                                                & 32                & 1.000                 & 0.000           & $\infty$
                                                & 32                & 1.000                 & 0.000           & $\infty$
                                                & 32                & 1.000                 & 0.000           & $\infty$           \\
    \midrule
    \multirow{5}{*}{Hyperprior}                 & 0.001
                                                & 0.141             & 505.530               & 119.795 & 61.536
                                                & 0.143             & 628.430               & 134.924 & 59.010
                                                & 0.206             & 0.300                 & 295.624 & 33.816
                                                & 0.140             & 0.200                 & 113.353 & 64.164
                                                & 0.219             & 0.100                 & 351.530 & 31.381                              \\
                                                & 0.004
                                                & 1.281             & 416.810               & 96.296 & 59.137
                                                & 1.305             & 443.720               & 111.149 & 56.154
                                                & 1.496             & 0.600                 & 270.809 & 31.951
                                                & 1.228             & 0.250                 & 89.969 & 60.669
                                                & 1.418             & 0.000                 & 325.956 & 28.875                              \\
                                                & 0.007
                                                & 1.802             & 483.240               & 90.191 & 56.089
                                                & 1.820             & 446.510               & 104.989 & 27.525
                                                & 1.972             & 0.500                 & 264.096 & 30.958
                                                & 1.744             & 0.250                 & 83.746 & 59.378
                                                & 1.957             & 0.000                 & 319.360 & 28.724                              \\
                                                & 0.010
                                                & 2.285             & 680.120               & 88.844 & 55.215
                                                & 2.329             & 689.010               & 103.464 & 18.409
                                                & 2.449             & 0.180                 & 261.373 & 11.876
                                                & 2.175             & 0.230                 & 82.161 & 56.790
                                                & 2.417             & 0.000                 & 316.723 & 6.536                              \\
                                                & 0.020
                                                & 3.217             & 356.390               & 84.905 & 19.667
                                                & 3.272             & 493.910               & 99.404 & 49.873
                                                & 3.550             & 0.300                 & 255.767 & 28.667
                                                & 3.069             & 0.270                 & 79.092 & 55.701
                                                & 3.543             & 0.100                 & 307.990 & 3.559                              \\
    \rowcolor{gray!20}
    \multicolumn{2}{l|}{Correlation}            &
    \multicolumn{4}{r|}{$\rho$ = +0.075}        &
    \multicolumn{4}{r|}{$\rho$ = +0.328}        &
    \multicolumn{4}{r|}{$\rho$ = +0.026}        &
    \multicolumn{4}{r|}{$\rho$ = -0.852}        &
    \multicolumn{4}{r}{$\rho$ = +0.301}                                                                                            \\
    \midrule
    \multirow{5}{*}{ELIC}                       & 0.001
                                                & 0.378             & 493.810               & 111.614 & 10.712
                                                & 0.382             & 642.070               & 126.456 & 10.824
                                                & 0.549             & 0.700                 & 282.581 & 4.795
                                                & 0.375             & 0.270                 & 105.355 & 12.425
                                                & 0.567             & 0.000                 & 338.763 & 4.139                              \\
                                                & 0.004
                                                & 1.477             & 363.310               & 89.766 & 11.396
                                                & 1.496             & 359.660               & 104.582 & 10.367
                                                & 1.864             & 0.800                 & 262.378 & 4.535
                                                & 1.413             & 0.200                 & 83.047 & 11.935
                                                & 1.765             & 0.100                 & 317.953 & 2.709                              \\
                                                & 0.007
                                                & 2.065             & 497.840               & 88.397 & 10.919
                                                & 2.090             & 479.770               & 103.119 & 10.037
                                                & 2.505             & 0.600                 & 255.803 & 4.449
                                                & 1.978             & 0.220                 & 80.661 & 11.505
                                                & 2.395             & 0.000                 & 313.257 & 3.895                              \\
                                                & 0.010
                                                & 2.528             & 410.190               & 86.314 & 10.722
                                                & 2.563             & 536.410               & 100.300 & 7.999
                                                & 2.735             & 0.200                 & 250.532 & 3.785
                                                & 2.429             & 0.190                 & 78.084 & 11.209
                                                & 2.670             & 0.400                 & 302.926 & 2.716                              \\
                                                & 0.020
                                                & 3.512             & 923.700               & 75.172 & 10.395
                                                & 3.565             & 1143.290              & 87.724 & 8.437
                                                & 3.824             & 0.120                 & 226.876 & 3.598
                                                & 3.325             & 0.300                 & 69.328 & 9.095
                                                & 3.702             & 0.800                 & 259.584 & 2.553                              \\
    \rowcolor{gray!20}
    \multicolumn{2}{l|}{Correlation}            &
    \multicolumn{4}{r|}{$\rho$ = -0.515}        &
    \multicolumn{4}{r|}{$\rho$ = -0.458}        &
    \multicolumn{4}{r|}{$\rho$ = +0.808}        &
    \multicolumn{4}{r|}{$\rho$ = +0.058}        &
    \multicolumn{4}{r}{$\rho$ = -0.940}                                                                                            \\
    \bottomrule
\end{tabular}
}
    \vspace{-2ex}
\end{table*}
\begin{table*}[t!] %
    \centering
    \begin{minipage}[t]{0.44\linewidth}
        \centering
        \caption{Rate-performance evaluation for CTTI (SD3.5).}
        \label{tab:sd35cond_rate_accuracy}
        \resizebox{\linewidth}{!}{

\begin{threeparttable}
    \begin{tabular}{ lc|rrrr|rrrr }
        \toprule
                                                             &
                                                             &
        \multicolumn{4}{c|}{Multimodal Conditioning Signals} &
        \multicolumn{4}{c}{Denoised Latents}                                                                                                                                                                                                       \\
                                                             & $\lambda$ & EBPFP$\downarrow$ & FID$\downarrow$         & MSE$\downarrow$ & B$_{\max}\uparrow$ & EBPFP$\downarrow$ & FID$\downarrow$         & MSE$\downarrow$ & B$_{\max}\uparrow$ \\
        \midrule
        Original                                             &           & 32                & 0.000\tnote{\ding{192}} & 0.000           & $\infty$           & 32                & 0.000\tnote{\ding{192}} & 0.000           & $\infty$           \\
        \midrule
        \multirow{5}{*}{Hyperprior}                          & 0.001     & 0.212             & 398.046                 & 8.225           & 47.598             & 0.106             & 295.636                 & 0.106           & 70.423             \\
                                                             & 0.004     & 0.895             & 252.742                 & 3.612           & 45.143             & 0.437             & 174.618                 & 0.057           & 69.016             \\
                                                             & 0.007     & 1.316             & 218.116                 & 2.515           & 45.934             & 0.755             & 142.491                 & 0.043           & 65.386             \\
                                                             & 0.010     & 1.495             & 245.572                 & 3.431           & 44.441             & 1.266             & 107.577                 & 0.032           & 67.078             \\
                                                             & 0.020     & 2.444             & 208.076                 & 4.867           & 42.380             & 2.131             & 66.029                  & 0.021           & 64.048             \\
        \rowcolor{gray!20}
        \multicolumn{2}{l|}{Correlation}                     &
        \multicolumn{4}{r|}{$\rho$ = +0.874}                 &
        \multicolumn{4}{r}{$\rho$ = +0.997}                                                                                                                                                                                                        \\
        \midrule
        \multirow{5}{*}{ELIC}                                & 0.001     & 0.627             & 323.198                 & 5.174           & 9.071              & 0.130             & 230.541                 & 0.080           & 14.151             \\
                                                             & 0.004     & 0.824             & 228.551                 & 2.534           & 8.897              & 0.449             & 138.303                 & 0.051           & 13.794             \\
                                                             & 0.007     & 1.309             & 228.456                 & 2.721           & 8.857              & 0.796             & 108.251                 & 0.039           & 13.843             \\
                                                             & 0.010     & 1.605             & 219.034                 & 5.316           & 8.715              & 1.087             & 92.164                  & 0.032           & 13.537             \\
                                                             & 0.020     & 2.165             & 193.243                 & 4.930           & 8.549              & 1.984             & 43.010                  & 0.018           & 13.162             \\
        \rowcolor{gray!20}
        \multicolumn{2}{l|}{Correlation}                     &
        \multicolumn{4}{r|}{$\rho$ = +0.216}                 &
        \multicolumn{4}{r}{$\rho$ = +0.999}                                                                                                                                                                                                        \\
        \bottomrule
    \end{tabular}

    \begin{tablenotes}
        \item[\ding{192}] We employ images generated from uncompressed features as the reference distribution. Consequently, the theoretical lower bound for FID is 0.
    \end{tablenotes}
\end{threeparttable}
}
    \end{minipage}
    \hfill
    \begin{minipage}[t]{0.545\linewidth}
        \centering
        \caption{Rate-performance evaluation for CVU (Layer 40 of DINOv3).}
        \label{tab:dinov3_layer39_rate_accuracy}
        \resizebox{\linewidth}{!}{
\begin{tabular}{
        rrrr
        |rrrr
        |rrrr}
    \toprule
    \multicolumn{4}{c|}{ImageNet-Val}    &
    \multicolumn{4}{c|}{ImageNet-A}      &
    \multicolumn{4}{c}{ImageNet-R}                                                                                                 \\
    EBPFP$\downarrow$                    & $\mathcal{A}\uparrow$ & MSE$\downarrow$       & B$_{\max}\uparrow$
                                         & EBPFP$\downarrow$     & $\mathcal{A}\uparrow$ & MSE$\downarrow$    & B$_{\max}\uparrow$
                                         & EBPFP$\downarrow$     & $\mathcal{A}\uparrow$ & MSE$\downarrow$    & B$_{\max}\uparrow$ \\
    \midrule
    32                                   & 1.000                 & 0.000                 & $\infty$
                                         & 32                    & 1.000                 & 0.000              & $\infty$
                                         & 32                    & 1.000                 & 0.000              & $\infty$           \\
    \midrule
    0.061                                & 0.400                 & 20934.370             & 170.715
                                         & 0.059                 & 0.400                 & 20184.156          & 171.689
                                         & 0.060                 & 0.570                 & 20297.203          & 169.217            \\
    0.409                                & 0.610                 & 3645.446              & 165.523
                                         & 0.375                 & 0.780                 & 3299.922           & 167.378
                                         & 0.398                 & 0.790                 & 3526.534           & 170.365            \\
    0.779                                & 0.720                 & 2430.534              & 163.823
                                         & 0.780                 & 0.840                 & 2055.999           & 162.902
                                         & 0.786                 & 0.920                 & 2279.332           & 161.597            \\
    0.886                                & 0.910                 & 2541.486              & 163.973
                                         & 0.852                 & 1.000                 & 2178.686           & 164.803
                                         & 0.863                 & 1.000                 & 2465.789           & 164.041            \\
    1.249                                & 0.980                 & 1754.262              & 158.072
                                         & 1.200                 & 1.000                 & 1442.532           & 159.318
                                         & 1.212                 & 1.000                 & 1649.976           & 160.048            \\
    \rowcolor{gray!20}
    \multicolumn{4}{r|}{$\rho$ = -0.818} &
    \multicolumn{4}{r|}{$\rho$ = -0.941} &
    \multicolumn{4}{r}{$\rho$ = -0.913}                                                                                            \\
    \midrule
    0.093                                & 0.320                 & 15768.326             & 52.907
                                         & 0.078                 & 0.370                 & 15859.486          & 53.098
                                         & 0.085                 & 0.470                 & 15971.028          & 53.173             \\
    0.420                                & 0.810                 & 3585.129              & 52.000
                                         & 0.361                 & 0.910                 & 2845.362           & 52.019
                                         & 0.387                 & 0.930                 & 3120.872           & 52.350             \\
    0.633                                & 0.910                 & 4592.171              & 51.728
                                         & 0.582                 & 0.980                 & 3297.489           & 51.798
                                         & 0.599                 & 0.980                 & 3445.326           & 51.463             \\
    0.954                                & 0.940                 & 3349.014              & 50.414
                                         & 0.821                 & 1.000                 & 2754.463           & 51.168
                                         & 0.868                 & 1.000                 & 3142.317           & 50.884             \\
    1.205                                & 0.980                 & 2057.627              & 50.205
                                         & 1.154                 & 1.000                 & 1936.007           & 50.449
                                         & 1.162                 & 1.000                 & 1953.272           & 50.406             \\
    \rowcolor{gray!20}
    \multicolumn{4}{r|}{$\rho$ = -0.979} &
    \multicolumn{4}{r|}{$\rho$ = -0.991} &
    \multicolumn{4}{r}{$\rho$ = -0.992}                                                                                            \\
    \bottomrule
\end{tabular}
}
    \end{minipage}
    \vspace{-2ex}
\end{table*}

\subsubsection{Impact of Split Points}
\label{sec:impact_of_split_points}

To evaluate the conventional strategy of detaching only the classification head~\cite{FeatureCoding-DA-UFC}, we compare DINOv3's deepest features (Layer 40, \cref{tab:dinov3_layer39_rate_accuracy}) against the early-stage split (Layer 10, \cref{tab:dinov3_rate_accuracy}).
This comparison reveals a behavioral divergence, particularly in the low-bitrate regime, stemming from the differences in representation characteristics across network depths.
Under severe compression (\eg $\lambda$=$0.001$), Layer 10 features suffer a functional collapse to near-zero accuracy.
Because these features contain dense, fine-grained low-level details, heavy compression irreparably corrupts their foundational spatial structure.
This initial error is then drastically amplified by subsequent non-linear blocks.
In contrast, coding deep features under identical constraints maintains a resilient performance floor ($0.32$-$0.57$).
Having already consolidated abstract semantics, these embeddings possess inherent robustness, preserving sufficient class-discriminative boundaries for the linear head to maintain partial functionality.
These contrasting results underscore the necessity of moving beyond one-size-fits-all feature coding.
To satisfy diverse edge-side constraints, codec design must evolve into a depth-adaptive paradigm: early-stage codecs must explicitly model and mitigate downstream error propagation, while deep-stage codecs can leverage semantic robustness to achieve extreme bandwidth reduction without risking catastrophic task failure.

\begin{table}[t!]
    \centering
    \caption{Peak GPU memory allocated (in MB) for encoding and decoding features across various feature sets with a batch size of 1, on a single NVIDIA GeForce RTX 4090 GPU.}
    \label{tab:gpu_memory}
    \resizebox{\linewidth}{!}{\begin{tabular}{llcccc}
    \toprule
    \multicolumn{2}{c}{\multirow{2}{*}{Feature Set}} &
    \multicolumn{2}{c}{Hyperprior}                   &
    \multicolumn{2}{c}{ELIC}                                                                                      \\
    \cmidrule(lr){3-4}
    \cmidrule(lr){5-6}
                                                     &                        & Encode & Decode & Encode & Decode \\
    \midrule
    \multirow{5}{*}{CVU (DINOv3)}                    & ImageNet-Val           & 1733   & 1754   & 2670   & 2674   \\
                                                     & ImageNet-A             & 1733   & 1754   & 2671   & 2674   \\
                                                     & ImageNet-R             & 1733   & 1754   & 2670   & 2674   \\
                                                     & ADE20K-Val             & 5341   & 5404   & 7848   & 7860   \\
                                                     & NYUDepthV2-Test        & 5233   & 5295   & 7695   & 7706   \\
    \midrule
    \multirow{5}{*}{CLU (Qwen3)}                     & GSM8K                  & 233 & 236 & 879 & 879  \\
                                                     & ArcChallenge           & 354 & 357 & 1071 & 1071  \\
                                                     & TruthfulQA             & 353 & 356 & 993 & 992  \\
                                                     & Hellaswag              & 624 & 627 & 1030 & 1028  \\
                                                     & Winogrande             & 230 & 232 & 949 & 949  \\
    \midrule
    \multirow{5}{*}{CLU (FalconMamba)}               & GSM8K                  & 232 & 234 &    657 & 654 \\ 
                                                     & ArcChallenge           & 354 & 357 &    680 & 678 \\ 
                                                     & TruthfulQA             & 365 & 367 &    693 & 691 \\ 
                                                     & Hellaswag              & 630 & 632 &    1043 & 1039 \\ 
                                                     & Winogrande             & 229 & 231 &    657 & 655 \\ 
    \midrule
    \multirow{5}{*}{CAU (KimiAudio)}                 & Librispeech-Test-Clean & 568    & 453    & 778    & 776    \\
                                                     & Librispeech-Test-Other & 390    & 392    & 689    & 687    \\
                                                     & AdvBench               & 203    & 205    & 414    & 423    \\
                                                     & OpenbookQA             & 470    & 472    & 807    & 804    \\
                                                     & SD-QA                  & 180    & 187    & 655    & 662    \\
    \midrule
    CTTI (SD3.5)                                     & COCO2017-Val (Caption) & 234    & 238    & 976    & 976    \\
    \bottomrule
\end{tabular}
}
\end{table}

\subsubsection{Generalizability Analysis}
\label{sec:generalizability_analysis}

We assess the generalizability of these universal baselines by analyzing their R-P trade-offs on entirely unseen model architectures, feature forms, and data modalities.
Originally trained by~\cite{FeatureCoding-DT-UFC} on features from DINOv2~\cite{oquab2023dinov2}, Llama3~\cite{dubey2024llama}, and Stable Diffusion 3~\cite{StableDiffusion3.5}, these codecs are evaluated on diverse new targets within our datasets, as summarized in~\cref{tab:dinov3_rate_accuracy,tab:qwen_rate_accuracy,tab:falconmamba_rate_accuracy,tab:kimiaudio_rate_accuracy,tab:sd35cond_rate_accuracy}.
First, regarding \textit{model architectures}, we investigate whether the learned codecs can generalize to an unseen state-space model.
As shown in~\cref{tab:falconmamba_rate_accuracy}, the codecs exhibit strong robustness to this architectural shift.
The resulting R-P trade-offs are highly comparable to those observed on transformer-based architectures (\cref{tab:dinov3_rate_accuracy,tab:qwen_rate_accuracy}), demonstrating an even more stable correlation between distortion and downstream performance across data subsets.
However, generalizing across fundamental data modalities poses a significant challenge, as evidenced by evaluations on CAU (\cref{tab:kimiaudio_rate_accuracy}) and CTTI (\cref{tab:sd35cond_rate_accuracy}).
When applying the codecs to KimiAudio's audio-text features, we observe severe performance degradation as shown in~\cref{tab:kimiaudio_rate_accuracy}.
Although the codecs achieve substantial bitrate reductions, downstream task metrics entirely collapse.
Similarly, as shown in~\cref{tab:sd35cond_rate_accuracy}, coding multi-modal conditioning signals results in persistently high FID scores across varying bitrates, indicating a severe deviation from the original prediction distribution.
These codecs exhibit robust intra-modal generalizability, effectively adapting to the unseen model architecture.
However, their failure on multi-modal features (\eg CAU and CTTI) underscores that developing universal \task methods capable of dynamically adapting to unseen complex semantic modalities remains a critical open challenge.

\subsubsection{Reconstructed Feature Analysis}
\label{sec:reconstructed_feature_analysis}

We visualize the histograms and cumulative distribution function (CDF) curves of the \textcolor{SteelBlue}{original} and \textcolor{DarkOrange}{reconstructed} features in~\cref{fig:histogram_cdf_distribution}.
While these baselines preserve the central mass of the distributions, as evidenced by the tightly overlapping CDF curves, they struggle at the tails when applied to large model representations.
The statistical illustrations highlight a clear truncation of outliers and a noticeable reduction in overall variance across most layers.
While robust architectures (\eg deep visual or language models) can tolerate mild outlier truncation at high bitrates (as evidenced by the high performance recovery at $\lambda=0.02$ in~\cref{tab:dinov3_rate_accuracy,tab:qwen_rate_accuracy,tab:falconmamba_rate_accuracy,tab:dinov3_layer39_rate_accuracy}), this statistical mismatch potentially impairs coding efficacy.
As bitrates decrease, or when applied to highly sensitive multi-modal features (\cref{tab:kimiaudio_rate_accuracy,tab:sd35cond_rate_accuracy}), the aggressive smoothing of these critical outliers~\cite{OutlierDimensions} leads to rapid performance degradation.
The visualizations also expose highly irregular internal distribution patterns that exacerbate compression bottlenecks.
Many features are highly sparse, forming sharp and discrete peaks rather than smooth and continuous curves.
Baseline methods tend to over-smooth these isolated peaks, which destroys fine-grained structural details.
Moreover, the internal dynamic ranges of these features are often wide and highly skewed.
Applying standard entropy models, originally designed for image priors, to process these heavy-tailed distributions inevitably leads to precision drops and information loss.
The structural mismatch in handling critical outliers and non-uniform spiky distributions reveals that directly applying image-centric coding frameworks to \task is suboptimal.
While they achieve high fidelity at high bitrates, they lack the distributional awareness required for optimal, stable compression across diverse and complex semantic modalities.
These limitations highlight the necessity of our dataset, which provides a necessary foundation for developing distribution-aware, specialized \task algorithms.

\subsubsection{Practicality Analysis}
\label{sec:practicality_analysis}

Existing studies mainly focus on the rate-performance trade-off, often overlooking deployment practicality in resource-constrained environments.
In practice, the computational latency and memory footprint of a codec can negate the theoretical benefits of data compression.
Regarding latency, a coding scheme is practical only if its end-to-end latency is less than the direct transmission time of the raw features.
Below this threshold, the codec provides a net latency advantage.
We define $\text{B}_{\max}$ as the maximum operating bandwidth where this holds true.
Let $T_{enc}$ and $T_{dec}$ denote the encoding and decoding time in feature coding, and $S_{raw}$ and $S_{enc}$ denote the size (in bits) of the raw and compressed features, respectively.
For a given net bandwidth $\text{B}$, the condition for a positive acceleration is:
\begin{equation}
    T_{enc} + \frac{S_{enc}}{\text{B}} + T_{dec} < \frac{S_{raw}}{\text{B}}
\end{equation}
Rearranging this inequality yields the condition $\text{B} < \text{B}_{\max}$:
\begin{equation}
    \text{B}_{\max} = \frac{S_{raw} - S_{enc}}{T_{enc} + T_{dec}}
\end{equation}
In this context, $\text{B}_{\max}$ acts as the \textbf{theoretical upper bound} on network bandwidth.
If the available bandwidth exceeds $\text{B}_{\max}$, the time saved by reducing the data volume is insufficient to offset the computational overhead of the codec.
The $\text{B}_{\max}$ values, measured in Mbps, are summarized in \cref{tab:dinov3_rate_accuracy,tab:qwen_rate_accuracy,tab:falconmamba_rate_accuracy,tab:kimiaudio_rate_accuracy,tab:sd35cond_rate_accuracy,tab:dinov3_layer39_rate_accuracy}.
Although ELIC achieves better rate-performance trade-offs compared to Hyperprior, this gain comes at the expense of practicality, reflected in a lower $\text{B}_{\max}$ and higher memory consumption (\cref{tab:gpu_memory}).
These findings highlight the need for future research to develop lightweight, high-throughput architectures that prioritize practical end-to-end latency and hardware efficiency over pure bitrate reduction.

\section{Limitations and Future Work}
\label{sec:limitation_and_featurework}

\subsection{Limitations of Current Benchmark}
\label{sec:limitations_of_current_benchmark}

While this benchmark provides a comprehensive foundational analysis of \task, it is subject to certain limitations.
Currently, following the practice of~\cite{FeatureCoding-DT-UFC,FeatureCoding-DA-UFC}, our evaluation focuses mainly on post-training compression, applying codecs to frozen large models.
We leave the exploration of joint codec-model optimization (\eg end-to-end fine-tuning or distillation) to future work.
Moreover, to establish the upper bound of resource requirements for actual codec execution, our practicality analysis profiles latency and memory primarily on desktop GPUs.
We leave further exploration on resource-constrained edge devices to future work, as real-world deployments often rely on specialized NPUs or DSPs, where the memory hierarchy and computational bottlenecks may exhibit substantially different behaviors.

\subsection{Towards Native Feature Coding}
\label{sec:towards_native_feature_coding}

Based on the systemic bottlenecks of existing feature codecs observed in our experiments, we identify three critical paradigm shifts required for the next generation of \task.

\parhead{Algorithmic Shift: Feature-Native and Distribution-Aware Coding.}
As analyzed in~\cref{sec:redundancy_analysis,sec:reconstructed_feature_analysis}, reshaping high-dimensional feature tensors into 2D grids (see~\cref{sec:feature_coding_pipeline}) imposes unreasonable geometric inductive biases, and standard quantization truncates critical heavy-tailed outliers.
Future coding pipelines must be \textbf{feature-native}, capable of modeling token-wise, spatial, or channel-wise redundancies directly without restricted 2D formats.
Furthermore, native codecs require distribution-aware entropy models that can accommodate the unbounded, spiky dynamic ranges of semantic embeddings without catastrophic information loss.

\parhead{Evaluation Shift: Semantics-Oriented Quality Assessment.}
As analyzed in~\cref{sec:distortion_performance_analysis}, this benchmark exposes the inadequacy of the conventional element-wise MSE metric in the context of \task, revealing a two-fold limitation: weak correlation with actual task utility and insensitivity to severe semantic drift.
The community needs new, semantics-oriented distortion metrics that align with task performance, enabling utility-driven rate-distortion optimization and accurately detecting the tipping points of task failure.
Ideally, such metrics should be differentiable to integrate into the pipeline, guiding codecs to adaptively allocate bits towards the most critical semantic dimensions.

\parhead{System Shift: Hardware-Algorithm Co-Design.}
The severe practicality bottleneck as revealed in~\cref{sec:practicality_analysis} dictates that pure bitrate reduction is meaningless if the computational overhead negates the transmission savings.
Furthermore, excessive memory footprint from the feature codec actively competes with the host model for limited resources.
On edge devices, this overhead can severely throttle actual model computation or even exceed hardware capacity, rendering the entire system inoperable.
Future research must transition from theoretical compression limits to practical, hardware-aware feature codecs that prioritize end-to-end latency, memory efficiency, and edge-device deployability.

\section{Conclusion}
\label{sec:conclusion}

In this paper, we formulate large model feature coding (\task) as a fundamental research problem for the efficient distributed deployment of modern large model systems.
To support systematic study, we establish \dataset, a comprehensive benchmark covering diverse task requirements, representative split points, and heterogeneous intermediate features.
Building on this benchmark, we introduce a unified evaluation protocol and examine representative universal codec baselines.
The results expose the broad limitations of the existing paradigm, revealing critical challenges across efficiency, distortion, efficacy, generalizability, and practicality.
These findings indicate that future \task research should move toward feature-native, semantics-aware, and hardware-efficient coding schemes.
We hope \dataset can provide a shared empirical basis for benchmarking, analyzing, and advancing future \task methods.

\appendix[Redundancy Analysis Metrics]
\label{sec:redundancy_analysis_metrics}

\subsection{Pearson Correlation Coefficient}
\label{sec:pearson_correlation_coefficient}

Given a packed feature map $\mathbf{X} \in \mathbb{R}^{H \times W}$, we compute $\rho$ along the horizontal ($h$) and vertical ($v$) axes to quantify local linear dependence.
To mitigate bias from dominant structures in specific rows or columns, we first compute the correlation between adjacent elements within each row (or column) and then average over valid rows (or columns) with non-zero variance.
Concretely, for the horizontal axis, let $\mathbf{x}_i = \mathbf{X}_{i,1:W-1}$ and $\mathbf{y}_i = \mathbf{X}_{i,2:W}$ be two length-$(W-1)$ vectors formed by a one-step left and right shift of the $i$-th row.
The horizontal correlation is
\begin{equation}
    \rho_h = \frac{1}{|\mathcal{I}_h|}\sum_{i\in\mathcal{I}_h}\rho(\mathbf{x}_i,\mathbf{y}_i)
    \quad
    \rho(\mathbf{x},\mathbf{y})=\frac{\mathrm{cov}(\mathbf{x},\mathbf{y})}{\sigma_{\mathbf{x}}\sigma_{\mathbf{y}}}
    \label{equ:pearson_correlation_coefficient}
\end{equation}
where $\mathcal{I}_h$ denotes the set of valid rows and $\sigma$ is the standard deviation.
The vertical correlation $\rho_v$ is computed analogously by correlating $\mathbf{X}_{1:H-1,j}$ with $\mathbf{X}_{2:H,j}$ for each column $j$ and averaging over valid columns.

\subsection{DCT Gini Coefficient \& Normalized Centroid}
\label{sec:dct_gini_coefficient_normalized_centroid}

To characterize spectral energy distribution, we apply a 2D DCT (type-II, orthonormal) to $\mathbf{X}$ and analyze the resulting coefficients $\mathbf{C}=\mathrm{DCT}(\mathbf{X})$.
We define the coefficient energies as
\begin{equation}
    e_{u,v} = |C_{u,v}|^2 \quad
    u=0,\dots,H-1 \quad
    v=0,\dots,W-1
\end{equation}
Let $\mathbf{e}=\{e_{(k)}\}_{k=1}^{N}$ be the energies $\{e_{u,v}\}$ flattened into a length-$N$ vector ($N=HW$) and sorted in non-decreasing order, \ie $e_{(1)}\le \cdots \le e_{(N)}$.
%
We measure energy sparsity using the Gini coefficient and define it as follows:
\begin{equation}
    G_{\text{DCT}} = \frac{\sum_{k=1}^{N}(2k-N-1)e_{(k)}}{N \sum_{k=1}^{N}e_{(k)}} \in [0,1]
    \label{equ:dct_gini_coefficient}
\end{equation}
In addition, we compute the normalized spectral centroid $C_{\text{DCT}}$ to capture where the energy is concentrated in the frequency plane.
Let $f_{u,v}$ denote the normalized radial frequency index at coefficient $(u,v)$, defined by the distance to the DC component:
\begin{equation}
    f_{u,v}=\frac{\sqrt{u^2+v^2}}{\sqrt{(H-1)^2+(W-1)^2}} \in [0,1]
    \label{equ:dct_frequency_index}
\end{equation}
The centroid is then computed as the energy-weighted average:
\begin{equation}
    C_{\text{DCT}} =
    \frac
    {\sum_{u=0}^{H-1}\sum_{v=0}^{W-1} f_{u,v} e_{u,v}}
    {\sum_{u=0}^{H-1}\sum_{v=0}^{W-1} e_{u,v}}
    \in [0,1]
    \label{equ:dct_centroid}
\end{equation}

{\small
\bibliographystyle{IEEEtran}
\bibliography{string2full,main_clean}
}
\end{document}